\documentclass[preprint,review,12pt]{elsarticle}

\usepackage{amssymb}

\usepackage{setspace}
\usepackage{caption}
\usepackage{amsmath}
\usepackage{systeme}
\usepackage{url}
\usepackage{xcolor}
\usepackage[linesnumbered,ruled,vlined]{algorithm2e}

\SetCommentSty{mycommfont}
\SetKwInput{KwInput}{Input}                
\SetKwInput{KwOutput}{Output}              

\journal{Pattern Recognition}

\begin{document}

\captionsetup[figure]{labelfont={bf},labelformat={default},labelsep=period,name={Fig.}}
\captionsetup[table]{labelfont={bf},labelformat={default},labelsep=period,name={Table}}
\begin{frontmatter}

\title{Refining a $k$-nearest neighbor graph for a computationally efficient spectral clustering}

\author{Mashaan Alshammari\corref{cor1}}
\ead{mals6571@uni.sydney.edu.au}
\author{John Stavrakakis}
\author{Masahiro Takatsuka}

\cortext[cor1]{Corresponding author. Tel.: +61 2 9351 3423; fax: +61 2 9351 3838}

\address{School of Computer Science, The University of Sydney, NSW 2006, Australia}

\begin{abstract}
\begin{singlespace}
Spectral clustering became a popular choice for data clustering for its ability of uncovering clusters of different shapes. However, it is not always preferable over other clustering methods due to its computational demands. One of the effective ways to bypass these computational demands is to perform spectral clustering on a subset of points (data representatives) then generalize the clustering outcome, this is known as approximate spectral clustering (ASC). ASC uses sampling or quantization to select data representatives. This makes it vulnerable to 1) performance inconsistency (since these methods have a random step either in initialization or training), 2) local statistics loss (because the pairwise similarities are extracted from data representatives instead of data points). We proposed a refined version of $k$-nearest neighbor graph, in which we keep data points and aggressively reduce number of edges for computational efficiency. Local statistics were exploited to keep the edges that do not violate the intra-cluster distances and nullify all other edges in the $k$-nearest neighbor graph. We also introduced an optional step to automatically select the number of clusters $C$. The proposed method was tested on synthetic and real datasets. Compared to ASC methods, the proposed method delivered a consistent performance despite significant reduction of edges.
\end{singlespace}
\end{abstract}


\end{frontmatter}




\section{Introduction}
\label{Introduction}

Spectral clustering gains popularity due to its ability of uncovering clusters with non-convex shapes. It uses the spectrum of pairwise similarity matrices to map data points to a space where they can be easily separated \cite{RN261,RN233,RN234,RN295}. Spectral clustering has been used in image segmentation \cite{RN228,RN296}, remote sensing image analysis \cite{RN297,RN276}, and detecting clusters in networks \cite{RN288,RN292,RN298,RN240}. Despite its elegance in uncovering clusters, spectral clustering comes with a heavy computational price. Decomposing the pairwise similarity matrix requires $\mathcal{O}(N^3)$ for $N$ data points. Spectral clustering is infeasible for applications with large $N$.

For the graph $G(V,E)$ represented by its affinity matrix $A$, reducing the size of $A$ means removing some vertices in $V$, whereas making $A$ sparser means removing edges. This is the motivation of approximate spectral clustering (ASC) \cite{RN296,RN297,RN276,RN232,RN275,RN285}, which adds two steps to the original algorithm of spectral clustering. First, it places $m$ prototypes in the data space where $m \ll N$. Then, spectral clustering is carried out on $m$ prototypes and uses the initial assignments to generalize the outcome. The $m$ prototypes are usually placed via vector quantization methods (e.g., $k$-means and self-organizing maps). Despite being a popular choice for approximate spectral clustering, vector quantization could converge badly, resulting in ill representation of data points due to randomness in initialization and/or training.

\begin{figure}[t]
	\centering
	\includegraphics[width=\textwidth,height=20cm,keepaspectratio]{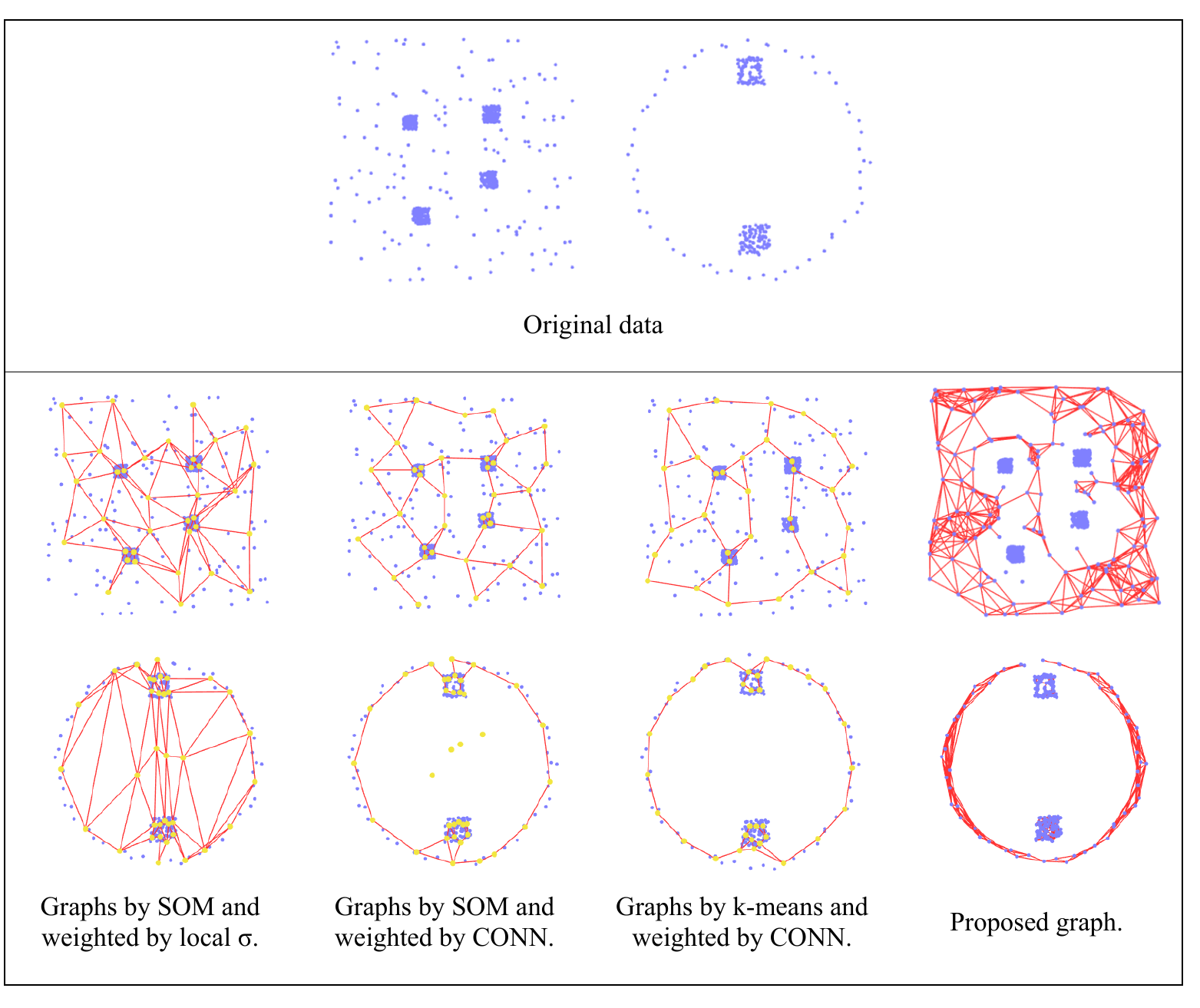}	
	\caption{Running approximate spectral clustering (ASC) via vector quantization on synthetic data (best viewed in color).}
	\label{Fig:Fig-01}
\end{figure}

Fig.\ \ref{Fig:Fig-01} illustrates two graphs by Zelnik-Manor and Perona \cite{RN237}, where approximate spectral clustering graphs were confused by the variation in local statistics. Generally, one can identify three deficiencies related to ASC via vector quantization: 1) these methods have a random selection step either in initialization or training which affects the consistency of clustering, 2) the obtained m prototypes provide a global overview of the data leaving out local information that could be crucial for clustering 3) vector quantization methods have to accommodate noisy data points as part of their training. Considering aforementioned deficiencies, we proposed a graph for spectral clustering $G=(V,E^\ast)$ where we kept the same number of vertices $V$ and find the most important subset of edges $E^\ast \subset E$. Our goal is to create a graph with less number of edges without compromising on the clustering accuracy. Hence, using a graph with edges less than $E^\ast$ would negatively impact the clustering accuracy.

To get $E^\ast$, we used a set of refinement stages that are computationally inexpensive. It starts by linking data points to their nearest neighbors progressively and monitor local statistics to stop linking when links appear outside the local neighborhood of points. Then, we used mutual k-nearest neighbor graph \cite{RN234} to filter edges that lack mutual agreement of the pair of points. The experiments show the proposed method outperforming ASC via vector quantization.

This paper is outlined as follows: in the next section we introduce spectral clustering. Also, we reviewed different approaches in the literature to construct a graph. In section 3, we present our approach for refining $k$-nn graph and automatically selecting the number of clusters $C$. In section 4, experiments were discussed.
\section{Spectral clustering (SC)}
\label{Spectralclustering}

The graph $G(V,E)$ connecting data points and its digital representation the affinity matrix A are the core components of spectral clustering. This clustering scheme is a relaxation of the normalized cut problem (Ncut) introduced by Shi and Malik \cite{RN261}. Their contribution made progress on the minimum cut (Mincut) defined as:
\begin{equation}
cut\left(B,\bar{B}\right)=\sum_{i\in B,j\in\bar{B}}\ A_{ij}\ ,
\label{Eq-001}
\end{equation}
where $B\subset\ V, \bar{B}$ is the complement of $B$, and $A_{ij}$ is the similarity score between the nodes $i$ and $j$. However, Mincut tends to cut isolated sets rather than significant partitions since it increases with the number of edges \cite{RN261,RN234}. Consequently, Ncut was introduced as:
\begin{equation}
Ncut\left(B,\bar{B}\right)=\frac{cut\left(B,\bar{B}\right)}{assoc\left(B,V\right)}+\frac{cut\left(B,\bar{B}\right)}{assoc\left(\bar{B},V\right)}\ .
\label{Eq-002}
\end{equation}

It penalizes the cut cost by the total connections from the nodes in B to all nodes V in the graph. Let $y$ be the exact solution of $Ncut \left(B,\bar{B}\right)$, with $y_i= 1$ if $i\ \epsilon\ B$, and $-1$ otherwise. Then $Ncut\left(B,\bar{B}\right)$ can be optimized as:
\begin{equation}
\min_x {Ncut\left(x\right)}=\min_y {\frac{y^T\left(D-A\right)y}{y^TDy}}\ ,
\label{Eq-003}
\end{equation}
where $D$ and $A$ are the degree and affinity matrices respectively.

To exactly solve Ncut, we have to look for two subsets with strong intra-connections and relatively weak weights between them, which was shown to be an NP-complete problem by Shi and Malik \cite{RN261}. However, by relaxing y to take real values it was shown by Shi and Malik \cite{RN362} that equation \ref{Eq-003} can be minimized by solving the generalized eigenvalue system:
\begin{equation}
\left(D-A\right)y=\lambda Dy\ .
\label{Eq-004}
\end{equation}
The second smallest eigenvalue $\lambda^L$ of the graph Laplacian $L=D-A$ and its corresponding eigenvector $v^L$, provide an approximation for solving Ncut \cite{RN234,RN279}. When there is a partitioning between $B$ and $\bar{B}$ such that:
\begin{equation}
v_i^L = \systeme*{\alpha \text{,} \quad i\in B,\beta \text{,} \quad i\in \bar{B}} ,
\label{Eq-005}
\end{equation}
Then $B$ and $\bar{B}$ becomes the optimal Ncut with a value of $Ncut\left(B,\bar{B}\right)=\lambda^L$ \cite{RN279}. $v^L$ is used to bipartition the graph then the following eigenvectors are used to partition the graph further.

\subsection{SC grouping algorithm}
\label{SCGroupingAlgorithm}

Clustering through graph Laplacian eigenvectors could be done iteratively (i.e., ordered by eigenvalues) or by constructing an embedding space using top eigenvectors. The latter approach is more convenient and a well-known method for embedding space clustering was introduced by Ng, Jordan and Weiss \cite{RN233}. They proposed a symmetric graph Laplacian $L_{sym}=D^{-1/2}AD^{-1/2}$ where $D$ and $A$ are degree and similarity matrices respectively. $L_{sym}$ top eigenvectors constitute an embedding space in which points that are strongly connected will fall close to each other making clusters detectable by $k$-means.

\subsection{Graph construction}
\label{GraphConstruction}

When it comes to spectral clustering, it is all about quantifying similarities. Ideally, points in the same cluster are linked by large weights so they can fall close in the embedding space. A naïve approach of assigning weights would be through Euclidean distance. However, this is not a practical choice, since it only considers first-order relationships. In first-order relationships edges are drawn based on information from pair of points only. A more practical approach would be considering second-order relationships where edges are drawn based on information from the neighbors. In the following subsections, we will go through some of the popular methods to construct a graph whose similarity matrix is fed into spectral clustering.

\subsubsection{Conventional graphs}
\label{ConventionalGraphs}

Conventional choices of constructing a graph include $k$-nearest neighbor graph and $\epsilon$-neighborhood graph. These graphs use first-order relationships. In nearest neighbor graph, each point is linked to $k$ points of its nearest neighbors. While in $\epsilon$-neighborhood graph, each point becomes a center of a sphere of radius $\epsilon$ and link with all points inside that sphere. These are straightforward approaches for constructing a graph, but their reliance on first-order relationships and hyperparameters limit their usability. Some restrictions could be applied to boost their performance. For example, connect with k-neighbors if they are closer than a threshold distance. Interested reader is referred to section 2.2 in \cite{RN234} and Appendix D in \cite{RN274}.

\subsubsection{Approximate graphs}
\label{ApproximateGraphs}

Approximate graphs use vector quantization method to construct a graph using a reduced set of prototypes. These methods can be classified into two categories: 1) methods that only places prototypes in the feature space such as $k$-means \cite{RN240,RN309}, 2) methods that are capable of placing prototypes and connecting them by edges (self-organizing map \cite{RN298,RN42} and neural gas \cite{RN259,RN257}). $k$-means attempts to minimize the sum of squared distances between points and their closest prototypes. Self-organizing map (SOM) uses a predefined lattice that connects prototypes. During SOM training, a winning prototype would pull its neighbors in the lattice towards the selected data point. Neural gas (NG) was an improvement over SOM since it links prototypes based on their location on the feature space not on the lattice. During NG training, the winning prototype would link to its closest neighbor, and that edge is allowed to age and “die” if it is not updated again. Both SOM and NG can produce a graph with less edges and vertices making spectral clustering computationally efficient.

Once the vector quantization training finishes, pairwise similarities could be set as: 1) a prototype to prototype similarity (approximate graph \cite{RN276,RN275})) or 2) a prototype to data point similarity (anchor graph \cite{RN359,RN358}). The former was used as a benchmark in experiments due to its a larger presence in the literature. Connectivity matrix (CONN) \cite{RN275} defines the similarity for a pair of prototypes according to the induced Delaunay triangulation \cite{RN257} which links the pair if there exists a data point that selects them as first and second best matching units (BMUs). When such a point does not exist, the pair of prototypes are not linked which makes CONN capable of producing sparse graphs. Growing neural gas (GNG) was used in \cite{RN296} as approximation graph. GNG applies the same training as NG, but in an incremental manner, where prototypes introduced to bridge the gaps during training. A comparison study \cite{RN285} discussed different approximate graphs and how to assign their weights using local scaling \cite{RN237} or CONN \cite{RN276,RN275}.

\subsubsection{Proximity graphs}
\label{ProximityGraphs}

In proximity graphs, a pair of points are linked if they satisfy a predefined condition. This makes them use second-order relationships since the linking decision is based on neighbors. In \cite{RN254}, that condition was if the neighborhood between pair of points is empty from any other point, then the pair should be linked. This type of graphs is known as empty region graphs (ERGs). ERGs rely on $\beta$ parameter to identify the neighborhood, that should be empty, to link the pair with an edge. Edges in the graph were locally scaled using similarity metric in \cite{RN237} to achieve accurate clustering. A more sophisticated condition for the empty region could be parameter free. Inkaya et al. introduced neighborhood construction NC algorithm \cite{RN303}. It starts by assigning direct neighbors as core neighbors and indirectly connects a point to other points through its core neighbors. Then it tracks the density between each pair, a pair having a density zero represents core neighbors. Once each point has a list of neighbors, the method tests the mutuality between neighbors’ lists, and drops the points lacking mutual agreement. Drawback of NC include isolated vertices, subgraphs, and asymmetric similarity matrix. This was rectified in \cite{RN253}, where additional steps were proposed to achieve symmetric similarity matrix. An undirected graph was constructed using NC, and if its connected components is less than the desired number of clusters $C$, edges were introduced between nearest points to satisfy the condition. Proximity graphs are capable of capturing underlying shape of the data. However, they came with a heavy computational price or the need for a hyperparameter. Density calculation alone requires $\mathcal{O}(n^3)$ \cite{RN303}.

\subsubsection{Constrained graphs}
\label{ConstrainedGraphs}

Constrained graphs require special types of constraints prior to data linkage. These constraints are must-link ML and cannot-link CL to force link or unlink of data points regardless of their location in the feature space. Authors of \cite{RN300} found that constraints are limited to a small number of points which is not very useful for clustering. They introduced an affinity propagation method where points are linked not only based on their affinity but with evaluation of nearby constraints. This makes a greater impact of constraints to improve clustering. Another approach introduced by Li, Liu and Tang \cite{RN302}. Instead of applying the constraints to the similarity matrix, they were applied to the eigenvectors constituting the embedding space. In that space must-link points should be close to each other and cannot-link points should be far apart. The authors created a measure of “good representation” that should hold the minimum cost when optimized. This method has a computational advantage over applying constraints to the similarity matrix. A separation of constrained graphs was proposed in \cite{RN301}. Must-link graph and cannot-link graph were created, and bi-objective graph optimization employed instead of eigen-decomposition. It is clear that constraining the graph created from data points would get better clustering results. However, this comes at a price of fundamentally changing the problem into semi-supervised instead of unsupervised. Constraints are usually created from the ground truth.
\section{Refined $k$-nearest neighbor graph for Spectral clustering}
\label{ProposedApproach}

From the previous overview, it can be noticed that there is no graph selection that works for different sets of data. Every method has to compromise at some stage, but we believe that local statistics between data points represent clustering information and should not be compromised. The ultimate goal of spectral clustering is to detect non-convex clusters, and local statistics are crucially important to achieve that goal. They have been avoided in approximate spectral clustering for computational efficiency. Also, most of the weirdly shaped clusters could be detected by approximation as long as they are dense, but this is not always the case.

\begin{figure}[t]
	\centering
	\includegraphics[width=\textwidth,height=20cm,keepaspectratio]{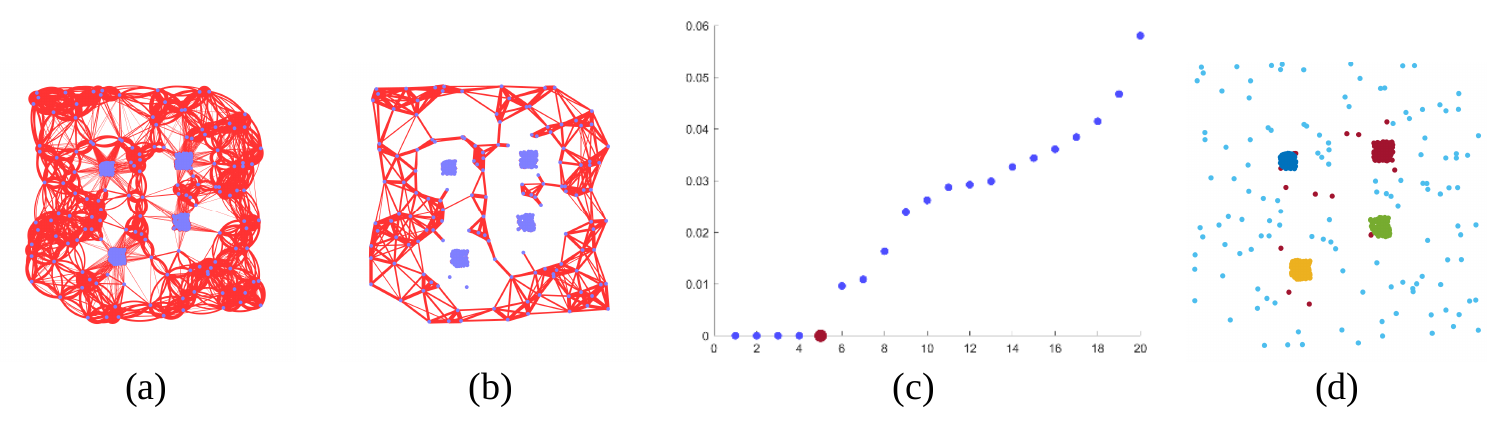}	
	\caption{A summary of the proposed approach. (a) $k$-nearest neighbor graph with $k$ set according to local statistics; (b) mutual $k$-nearest neighbor graph to filter edges lacking mutual agreement; (c) an optional step to locally monitor the change in eigenvalues to detect the number of clusters $C$; (d) clustering outcome (best viewed in color).}
	\label{Fig:Fig-02}
\end{figure}

Our method attempts to balance the tradeoff between locally scaled graphs and computational efficiency. It starts by creating $k$-nearest neighbor graph at each point and stop when it violates local statistics. Then, a mutuality check was run to ensure agreement among data points. We also introduced an eigengap detection method to uncover number of clusters $C$. Overview of the method is shown in Fig.\ \ref{Fig:Fig-02}.

\subsection{Setting $k$ in k-nearest neighbor graph}
\label{ProposedApproach-1}

Conventional $k$-nearest neighbor graphs have the problem of treating all data points equally. Due to their location in feature space, data points have different needs for the number of edges that could be larger or smaller than $k$. Ignoring these needs, and forcing each point to have $k$ edges, might result in linking two clusters or breaking a single cluster. Therefore, the parameter $k$ should be adaptively computed to accommodate the needs for each data point.

\begin{figure}[t]
	\centering
	\includegraphics[width=0.79\textwidth,height=20cm,keepaspectratio]{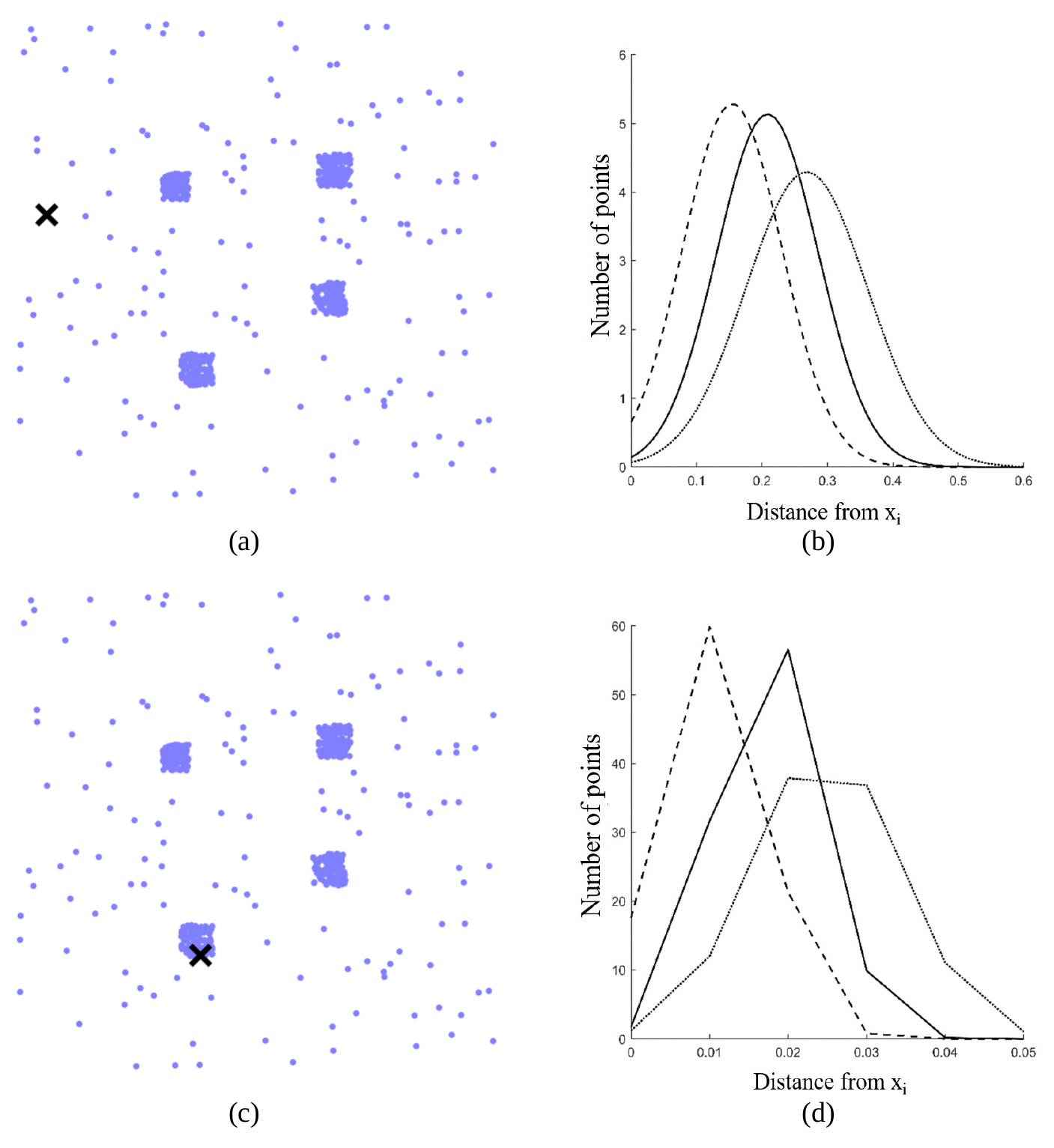}	
	\caption{Distribution of distances from the point marked as $\times$. (dashed line) $k=7$ (solid line) $k$ set by the proposed method (dotted line) $k$ set by the proposed method plus 20 neighbors.}
	\label{Fig:Fig-03}
\end{figure}

Setting $k$ manually for each data point is not a practical process. Therefore, we used the distribution of distances to monitor how it changes as we add more neighboring distances to the distribution. The intuition behind it is at some point we are leaving the current cluster to another cluster while adding more neighbors. This movement between clusters should be reflected on the distribution of distances. First, we need a baseline distribution of distances indicating how distances are distributed in the current cluster. Also, we need a threshold to notify us that we are leaving the baseline distribution towards something different. Our baseline distribution is the normal distribution of distances to the first seven neighbors, this was selected for locally scaled pairwise distances in \cite{RN237,RN274}. By setting the baseline distribution to seven we’re implying that the first seven neighbors belong to the same cluster. This is a strong assumption and it fails in large datasets (as we will see in Table \ref{Table:Table-03}) where we had to change it to be fifty, because seven was very narrow. that’s why we are currently working on figuring this parameter automatically.

Since we are moving in one direction which is far from the mean $\mu$ of the baseline distribution, we set the threshold to stop adding more neighbors as $ \lbrack \mu + \sigma \rbrack $.  Once the new mean is outside the interval $ \lbrack \mu + \sigma \rbrack $ we set the $k$ for the current point at its current neighbor. In Fig.\ \ref{Fig:Fig-03}, we compare the distribution of distances for two points sitting in different locations in the feature space. The plot on the right shows a dashed line that is a baseline distribution of distances $k=7$, solid line is the selected distribution set not to violate the interval $ \lbrack \mu + \sigma \rbrack $, dotted line is the distribution if we add 20 more neighbors beyond $ \lbrack \mu + \sigma \rbrack $. The selected distribution kept the shape of baseline distribution while adding more edges that are useful for clustering. However, if we keep adding more neighbors beyond$ \lbrack \mu + \sigma \rbrack $ interval, the distribution becomes flatter losing the shape of the baseline. Interestingly the distribution of distances indicates the sparsity of a cluster. In Fig.\ \ref{Fig:Fig-03} (a), the point marked as $\times$ sits in a sparse cluster making its distribution more naturally looking covering large interval (0.1,0.4), due to varying distances around it. On the other hand, the point in bottom row sitting in a dense cluster has a sharp distribution covering a narrow interval (0.01,0.04), given similar distances around it.

\begin{algorithm}[t]
	\DontPrintSemicolon
	
	\KwInput{$N$ data points, maximum number of neighbors $k_{max}$}
	\KwOutput{Refined $k$-nearest neighbor graph}
	
	\tcc{The following step has computations in order of $\mathcal{O}(dNlogN)$}
	Construct $k$-nn graph where $k=k_{max}$ represented by its distance matrix $D(N,k_{max})$
	
	\tcc{The following loop has computations in order of $\mathcal{O}(N_{kmax})$}
	\For{$i = 1 \text{ to } N$}
	{
		\For{$j = 1 \text{ to } k_{max}$}
		{
			$DM_{i,j}$ = mean($D_{i,\ 1\ to\ j}$) + standard deviation($D_{i,\ 1\ to\ j}$)
		}
	}

	let $DM7(N,k_{max})$ be an empty matrix
	
	let all columns in $DM7$ equal the 7$^\text{th}$ column in $DM$
	
	$D^\ast(N,k_{max})=DM7(N,k_{max})-DM(N,k_{max})$
	
	\tcc{The following loop has computations in order of $\mathcal{O}(N_{kmax})$}
	\For{all elements in $D^\ast(N,k_{max})$}
	{
		\If{$D_{i,j}^\ast<0$}
		{
			$D_{i,j}^\ast=0$
		}
	}

	\tcc{The following loop has computations in order of $\mathcal{O}(N_{kmax})$}
	\For{all elements in $D^\ast(N,k_{max})$}
	{
		\If{$D_{i,j}^\ast==0$ or $D_{j,i}^\ast==0$}
		{
			$D_{i,j}^\ast=0$
			$D_{j,i}^\ast=0$
		}
	}

	Construct refined $k$-nn using distance matrix $D^\ast(N,k_{max})$

	\caption{Constructing a refined $k$-nearest neighbor graph}
	\label{Alg:Alg-01}
\end{algorithm}

The computational bottleneck is to get the initial $k$-nearest neighbor graph. These computations could be reduced by using efficient data structure like $kd$-trees which can be constructed in $\mathcal{O}(dNlogN)$ [34]. The parameter $k$ was set as $k_{max}$. By setting $k_{max}$ we are comfortable that each data point requires edges less than $k_{max}$. Then, getting the refined $k$-nearest neighbor graph requires low computations $\mathcal{O}(Nk_{max})$. Once we have the distance matrix of size $N\times\ k_{max}$, we compute mean and standard deviation $\mu_0$ and $\sigma_0$ for the first seven columns. Then we add more neighbors to compute the new mean $\mu_i$. Once $\mu_i>\mu_0 + \sigma_0$, the comparison stops and all elements up to $k_{max}$ are nullified. This process is illustrated by steps 1--5 in Algorithm \ref{Alg:Alg-01}.

\subsection{Checking mutual agreement}
\label{ProposedApproach-2}

\begin{figure}[t]
	\centering
	\includegraphics[width=\textwidth,height=20cm,keepaspectratio]{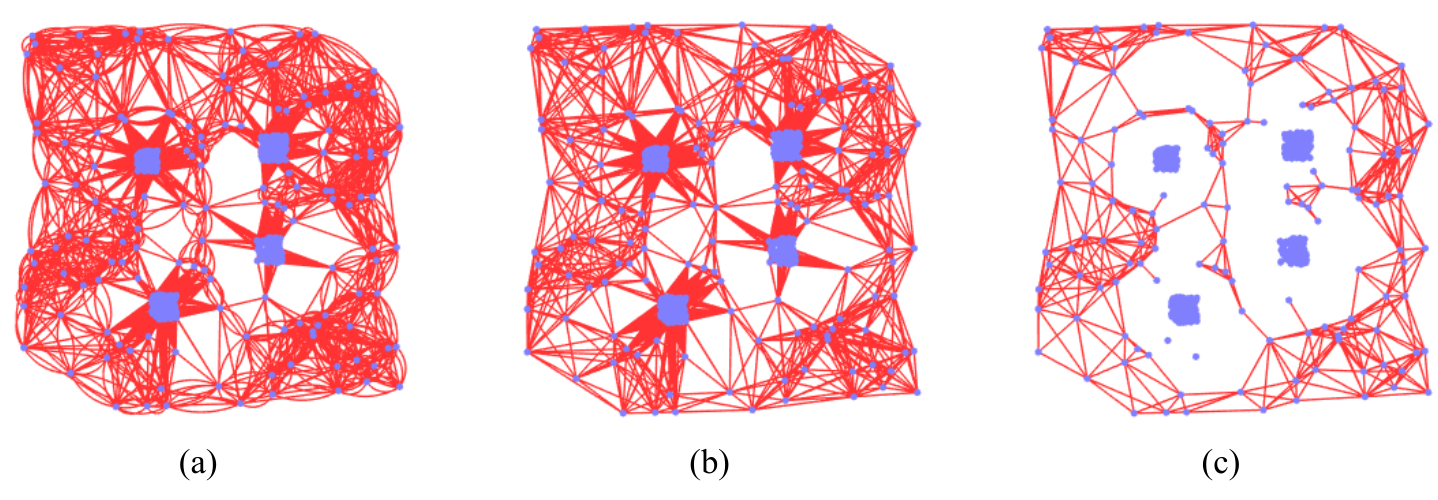}	
	\caption{Illustration on the importance of mutuality test. (a) refined $k$-nn graph with directed edges, (b) refined $k$-nn graph with undirected edges, (c) refined $k$-nn after removing edges lacking mutual agreement (best viewed in color).}
	\label{Fig:Fig-04}
\end{figure}

The graph obtained in the previous step is a directed graph. Each edge indicates the existence of the destination point in the source point refined $k$-nn list. A pair of points have a mutual agreement if they have each other in their refined $k$-nn lists. Fig.\ \ref{Fig:Fig-04} shows how crucial this step is. If we convert the refined $k$-nn graph Fig.\ \ref{Fig:Fig-04} (a) into undirected graph Fig.\ \ref{Fig:Fig-04} (b) and proceed with spectral clustering, we should not expect great results since all clusters are connected. However, if we drop the edges between neighbors lacking mutual agreement, we end up with a graph highlighting groups of points separated by different local statistics (check steps 6--7 in Algorithm \ref{Alg:Alg-01}).

\subsection{Looking for the eigengap}
\label{ProposedApproach-3}

Once the graph is ready, we can construct the pairwise similarity matrix and assign weights on edges. We used the similarity introduced by Zelnik-Manor and Perona \cite{RN237} it is superior in highlighting clusters with different statistics. The affinity matrix $A$ is defined as follows:
\begin{equation}
A_{ij}=\exp{\left(\frac{-d^2\left(i,j\right)}{\sigma_i\ \sigma_j}\right)}.
\label{Eq-006}
\end{equation}
The local scale $\sigma_i$ set as the distance of a point to its k$^\text{th}$ neighbor, here we set it to be the 7$^\text{th}$ neighbor $d(i,i_{k=7})$. The eigen-decomposition is carried out afterwards to get the top $C$ eigenvectors ($C$ is number of clusters). Setting $C$ manually is not practical. Our recent work in \cite{RN296} introduced a framework to uncover $C$ by evaluating eigenvectors independently and calculate Davies-Bouldin index (DBI) for each eigenvector. However, that framework was built around approximate spectral clustering where number of prototypes is $m$ compared to $N$ points in our case $m \ll N$. Applying the same framework here could be computationally expensive. Therefore, we preferred to track the change in eigenvalues. Liu et al. proposed predicting $C$ from eigenvalues, but their method requires a parameter $\tau$ \cite{RN168}. To keep it simple, we start with 2$^\text{nd}$ and 3$^\text{rd}$ eigenvalues and compute their mean and standard deviation. As we add more eigenvalues we check if the new mean is less than old mean plus standard deviation (see steps 1--3 in Algorithm \ref{Alg:Alg-02}). $C$ would be set as:
\begin{equation}
C=i,\ \ \ \ \ \ \ \text{where}\ \mu_{i+1}>\mu_i+\sigma_i.
\label{Eq-007}
\end{equation}

\subsection{Clustering in the embedding space}
\label{ProposedApproach-4}

\begin{figure}[t]
	\centering
	\includegraphics[width=\textwidth,height=20cm,keepaspectratio]{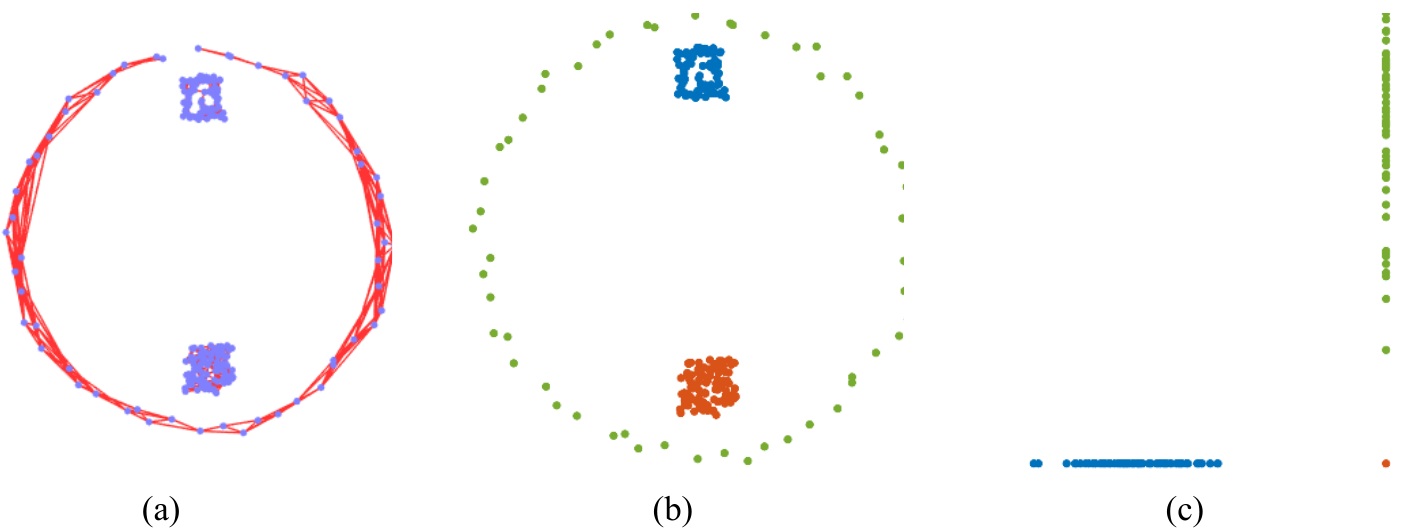}	
	\caption{Illustration on the importance of mutuality test. (a) refined $k$-nn graph with directed edges, (b) refined $k$-nn graph with undirected edges, (c) refined $k$-nn after removing edges lacking mutual agreement (best viewed in color).}
	\label{Fig:Fig-05}
\end{figure}

In spectral clustering, it is not enough to specify only the number of clusters $C$, it also needs the number of dimensions of the embedding space. The original algorithm \cite{RN233} states that for $C$ clusters $k$-means should operates in the top $C$ eigenvectors space. In practice, $C$ eigenvectors could be detected in a space where number of eigenvectors is less than $C$. For example, in Fig.\ \ref{Fig:Fig-05} the original data points form three clusters, and by plotting them using top two eigenvectors, it is clear that clusters are separated and detectable via $k$-means. Therefore, it is worth checking how $k$-means would perform in an embedding space with less than $C$ eigenvectors.

\begin{algorithm}[!h]
	\DontPrintSemicolon
	
	\KwInput{A refined $k$-nn graph $G(V,E)$ with $N$ vertices,
		\newline number of required eigenvalues $\lambda_{max}$}
	\KwOutput{$N$ data points grouped into $C$ clusters}
	
	Compute graph Laplacian $L_{sym}=D^{-1/2}AD^{-1/2}$
	
	\tcc{The following step has computations in order of $\mathcal{O}(N^3)$}
	Compute eigenvalues $\lambda$ and eigenvectors $v$ of graph Laplacian $L_{sym}$
	
	\tcc{The following loop has computations in order of $\mathcal{O}(\lambda_{max})$}
	\For{$i = 3$ to $\lambda_{max}$}
	{
		\If{$\lambda_{i+1} > (\text{mean}(\lambda_{2 \text{ to } i}) + \text{standard deviation}(\lambda_{2 \text{ to } i})$}
		{
			set $C = i$
		}
	}
	
	\tcc{The following loop has computations in order of $\mathcal{O}(CdtN)$}
	\For{all elements in $D^\ast(N,k_{max})$}
	{
		run $k$-means with $k=C$ on $N$ data points using eigenvectors $v_2$ to $v_i$ of $v$.
		
		set $l_i$ as the label vector returned by $k$-means.
	}
	
	\tcc{The following loop has computations in order of $\mathcal{O}(CE)$}
	\For{$i = 2$ to $C$}
	{
		let $a_i$ be the variable holding sum of weights connecting unmatched labels in $l_i$.
		
		label all vertices in $V$ using $l_i$.
		
		\For{each edge $E(p,q)$ in $E$ where $p,q\in V$}
		{
			\If{labels are different $l_i(p)\neq l_i(q)$}
			{
				$a_i=a_i+E(p,q)$
			}
		}
	}
	
	Return the lowest $a_i$ and its associated $l_i$	
	
	\caption{Clustering in embedding space with unknown $C$}
	\label{Alg:Alg-02}
\end{algorithm}

For all eigenvectors less than $C$, we constructed embedding spaces starting by 2$^\text{nd}$ and 3$^\text{rd}$ eigenvectors then add one more eigenvector up to $C$. On each embedding space, $k$-means operates to detect $C$ clusters. We end up with $C$ vectors each of which represents cluster memberships. To choose the right cluster membership, we applied each membership vector to the refined $k$-nn graph and sum the weights of inter-cluster edges. Inter-cluster edges link points with different cluster membership. The right membership vector is the one that yields the lowest sum of inter-cluster weights. Ideally, this sum would be zero indicating no edges are linking different clusters (see steps 4--6 in Algorithm \ref{Alg:Alg-02}).

\subsection{Integration with SpectralNet}
\label{ProposedApproach-5}

Spectral clustering using deep neural networks (SpectralNet) was introduced by \citet{RN360}. They highlighted two shortcomings of spectral clustering: 1) the scalability issue with large datasets where direct computation of eigenvectors could be infeasible, and 2) the generalization issue which is extending spectral embedding to unseen data in a task commonly known as out-of-sample-extension \cite{Bengio2004OOSE,RN227,Coifman2006Geometric}.

Our proposed method could be integrated with SpectralNet. The SpectralNet consists of three main stages: 1) unsupervised learning of an affinity matrix given a distance measure, via a Siamese network, 2) unsupervised learning of an embedding space by optimizing spectral clustering objective, and 3) learning cluster assignments by running $k$-means in the embedding space. Our method for filtering the graph could be executed before running the Siamese net. Siamese nets \cite{Hadsell2006Dimensionality,Shaham2018Learning} are trained to learn complex affinity relations that cannot be captured by Euclidean distance. \citet{RN360} empirically found that using Siamese net to determine the affinity often improves the quality of clustering.

Siamese nets are usually trained on similar (positive) and dissimilar (negative) pairs of data points. For labeled data, positive and negative pairs could be decided from the labels. For example, a pair with the same label is set a positive pair, while a pair with different labels is set as negative pair. But this is not the case with unlabeled data where nearest neighbor graph can be used to determine positive and negative pairs. \citet{RN360} constructed positive pairs for Siamese net by pairing each point with two of its nearest neighbors. An equal number of negative pairs was randomly chosen from farther neighbors. We used a different approach in our experiments. We let our method detailed in Algorithm \ref{Alg:Alg-01} to decide how many neighbors a data points should have as positive pairs. Then, an equal number of farther neighbors is set as negative pairs.

Our approach to pass positive and negative pairs to Siamese has two advantages. First, the number of positive pairs is not fixed for all data points. This makes points in dense regions to have more positive and negative pairs. Second, we did not use a random selection for negative pairs, instead we assigned farther neighbors as negative pairs. This would contribute to the consistency of the method over independent executions.

\section{Experiments and discussions}
\label{Experiments}

Experiments were conducted using synthetic data, real data, and a dataset with an increasing noise (10\% to 50\% of $N$). The proposed method was compared against approximate spectral clustering (ASC) methods. The most famous vector quantization for ASC are: $k$-means and self-organizing map (SOM). These were selected for approximation. Similarities between prototypes obtained through $k$-means and SOM were computed using local $\sigma$ \cite{RN237}, CONN \cite{RN275}, and CONNHybrid \cite{RN276}. There were other similarity metrics proposed in \cite{RN276}, but they were built on top of CONN and their performance was highly correlating with CONN. All experiments were coded in MATLAB 2018b and run on a windows 10 machine (3.40 GHz CPU and 8 GB of memory). The code is available on \url{https://github.com/mashaan14/Spectral-Clustering}.

\subsection{Evaluation metrics}
\label{Experiments-1}

The performance of competing methods was evaluated by comparing labels obtained by the clustering method with the true labels provided in ground truth. Two metrics were used for the evaluation: clustering accuracy (ACC) and adjusted Rand index (ARI). Clustering accuracy checks one on one assignments and computes the percentage of hits. Let $T_i$ and $L_i$ be ground truth labels and labels obtained through clustering respectively. Then, accuracy is defined as \cite{RN238}:
\begin{equation}
ACC(T,L)=\frac{\sum_{i=1}^{N}{\ \delta(T_i,map(L_i))}}{N},
\label{Eq-008}
\end{equation}
where $N$ is the number of points and the function $\delta(x,y)$ equals one when $x=y$ and equals zero otherwise. The function $map(L_i)$ is the permutation mapping that maps the obtained cluster labels to its equivalent in the ground truth labels.

The adjusted Rand index (ARI) \cite{RN365} is one of the “pair counting evaluation measures”. For two groupings $T$ and $L$, ARI counts how many pairs $T$ and $L$ agreed or disagreed. It has better bounds than the original Rand index (RI) \cite{RN364}. The upper bound is 1 indicating identical groupings and the lower bound 0 indicates random groupings. Let $N$ be the number of elements in the contingency table with $T$ rows and $L$ columns. Given all possible pairs in $\binom{n}{2}$, they can be classified into four types: $n_{11}$: pairs in the same cluster in both $T$ and $L$; $n_{00}$: pairs in different clusters in both $T$ and $L$; $n_{01}$: pairs in the same cluster in $T$ but in different clusters in $L$; $n_{10}$: pairs in different clusters in $T$ but in the same cluster in $L$. Then, ARI is defined as:
\begin{equation}
ARI(T,L)=\frac{2(n_{00}n_{11}-n_{01}n_{10})}{(n_{00}+n_{01})(n_{01}+n_{11})+(n_{00}+n_{10})(n_{10}+n_{11})}\ .
\label{Eq-009}
\end{equation}

The computational efficiency of competing methods was measured by the percentage of edges used compared to all edges in a fully connected graph. This is more suitable measure than simply measuring the running time which is sensitive to the experimental setup (e.g., computation power, machine used, etc.). The metric is computed as follows:
\begin{equation}
E\%=\frac{E(G)}{E(G_{full})}\ ,
\label{Eq-009-01}
\end{equation}
where $E(G)$ is the number of edges in the filtered graph and $E(G_{full})$ is the number of edges in the fully connected graph.

\subsection{Synthetic datasets}
\label{Experiments-2}

\begin{figure}[t]
	\centering
	\includegraphics[width=\textwidth,height=20cm,keepaspectratio]{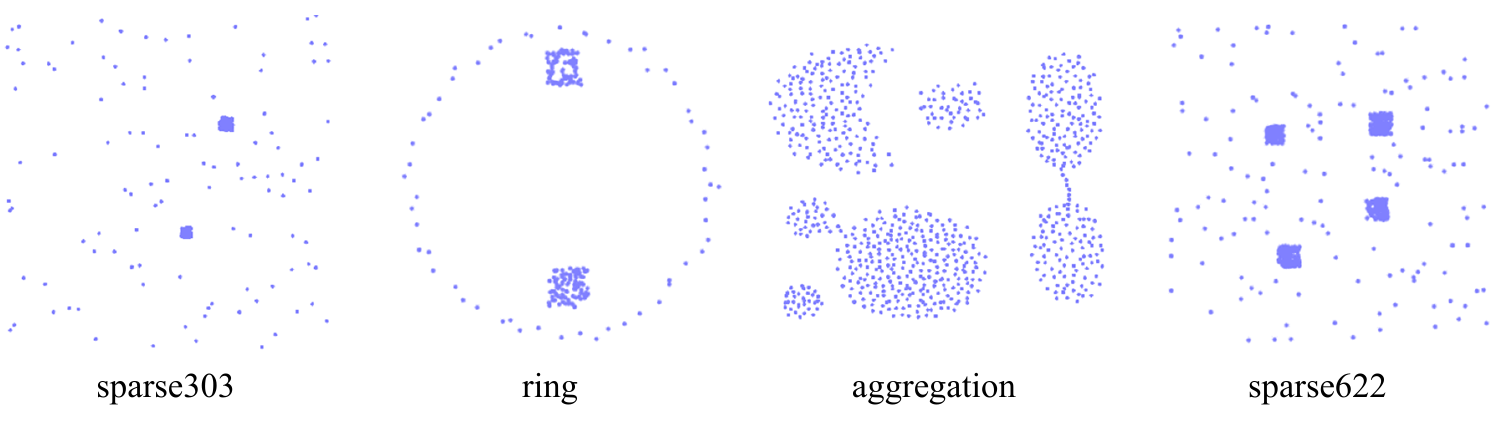}	
	\caption{Synthetic datasets used in the experiments.}
	\label{Fig:Fig-06}
\end{figure}

\begin{table}
	\centering
	\caption{Properties of tested datasets. $N =$ number of samples, $d =$ number of dimensions (i.e., features), and $C =$ number of clusters.}
	\includegraphics[width=\textwidth,height=20cm,keepaspectratio]{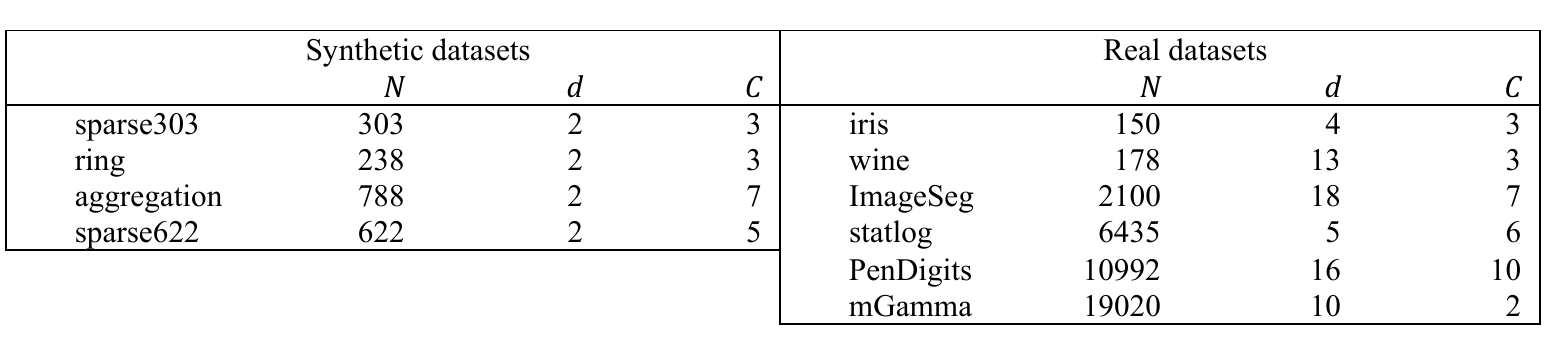}	
	\label{Table:Table-01}
\end{table}

For approximate spectral clustering (ASC), the number of prototypes was selected as the elbow point in the quantization error curve by $k$-means. It was set as 32, 32, 64, and 32 for \texttt{sparse303}, \texttt{ring}, \texttt{aggregation}, and \texttt{sparse622} respectively. As a general observation, prototypes placed by self-organizing map (SOM) performed better than the ones placed by $k$-means (especially in \texttt{sparse303} and \texttt{sparse622}). This behaviour could be explained by training of SOM in which prototypes move as a group towards a data point and not independent from each other as in $k$-means. This enables SOM to place more neurons into dense regions than $k$-means (please refer to Fig.\ \ref{Fig:Fig-01} for illustration). In terms of the similarity metric, CONN and CONNHybrid have a slight advantage over local $\sigma$ especially in \texttt{aggregation} dataset in which all clusters are dense. In \texttt{ring} dataset, CONN and CONNHybrid achieved similar performances higher than local $\sigma$.

\begin{table}[t]
	\centering
	\caption{Evaluating competing methods on synthetic data for 100 runs. Three rows for each dataset: \texttt{ACC}: mean accuracy $\pm$ standard deviation, \texttt{ARI}: mean adjusted Rand index $\pm$ standard deviation, and \texttt{E\%}: the percentage of edges compared to edges in a fully connected graph. Bold values are the best scores.}
	\includegraphics[width=0.94\textwidth,height=20cm,keepaspectratio]{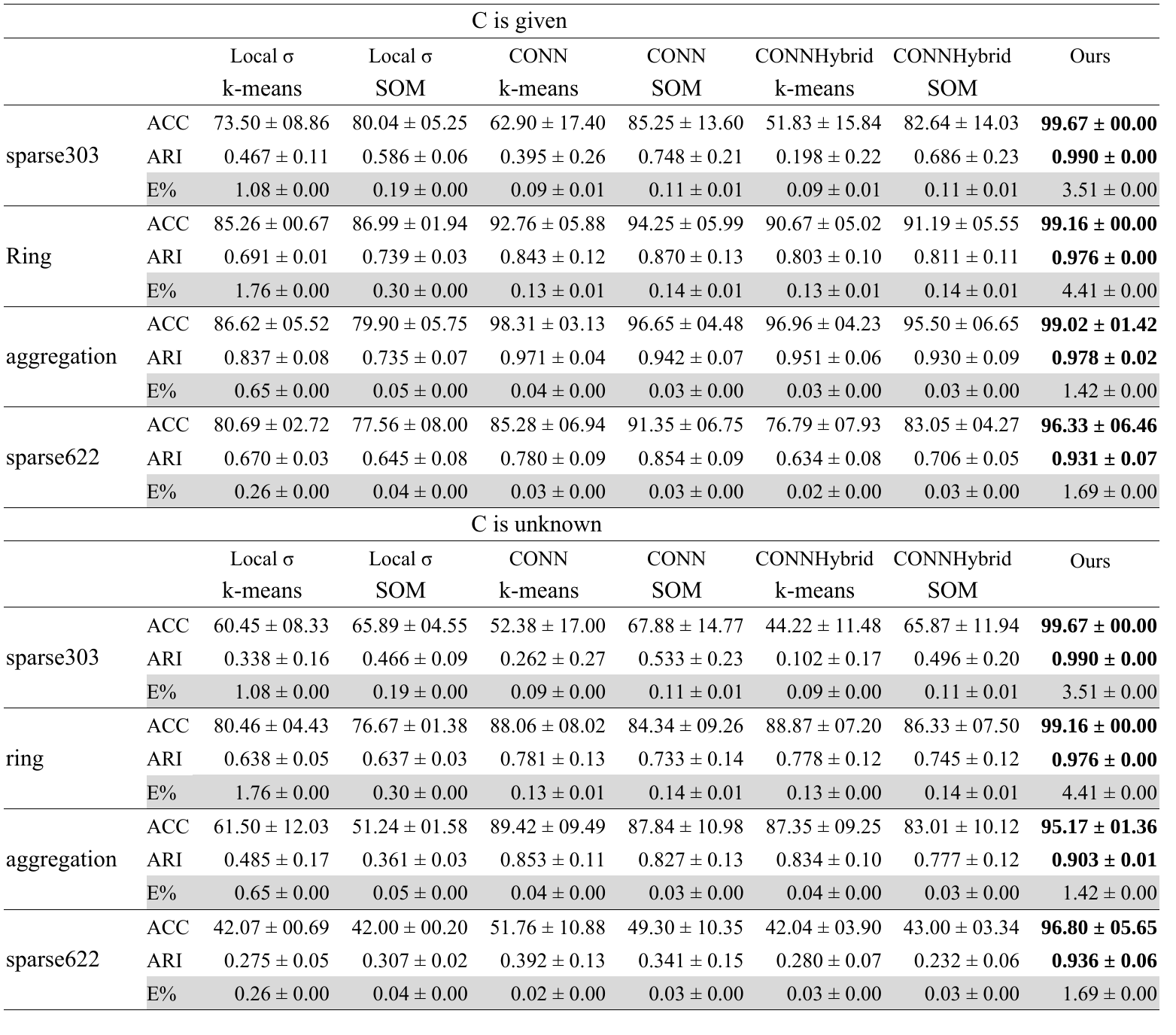}	
	\label{Table:Table-02}
\end{table}

There were performance drops when $C$ was unknown compared when it was given. The most significant drops occurred in \texttt{sparse622} when it reached 50\% drop compared to when $C$ was given. In \texttt{sparse303}, the drop was around 25\%. For \texttt{ring} and \texttt{aggregation} datasets the drop was smaller compared to sparse datasets with exception of local $\sigma$ in \texttt{aggregation} dataset. Usually in sparse datasets, the graph has a single connected component and by giving $C$ we are forcing the method to break it into the desired number of connected components. But when $C$ is unknown it was very difficult for the eigengap detection to uncover $C$ because it was a single connected component which cause the sharp drops in performance. For \texttt{ring} and \texttt{aggregation} datasets, ASC passes a graph with multiple connected components making the task easier for the eigengap detector.

The major drawback of ASC that kept persisting across all synthetic datasets is the standard deviation of performance scores. In different experiments it reached 17\% of ACC and 0.2 of ARI, this highlights the inconsistency of these methods. Our proposed method achieved a score close to a full mark across all synthetic datasets. Another advantage of our method was the consistent performance. This indicates that no matter how many times you repeat the method there is a high chance to get a score close to the full mark.

The third evaluation metric was the percentage of used edges compared to fully connected undirected graph. All ASC methods had a lower number of edges than our method because they are constructing a graph out of $m$ prototypes and ours used all data points $N$ ($m \ll N$). However, the highest percentage of edges for our method was 4.41\% in \texttt{ring} dataset. This means that 95.59\% were removed from the fully connected graph, this is a considerable reduction in computations and memory footprint.

\subsection{Real datasets}
\label{Experiments-3}

For real datasets the number of prototypes in ASC methods was set by monitoring quantization error to 32, 32, 40, 100, 500, and 1000 for datasets: \texttt{iris}, \texttt{wine}, \texttt{ImageSeg}, \texttt{statlog}, \texttt{Pen digits}, and \texttt{mGamma}. For \texttt{iris} dataset and when $C$ was given our method achieved the highest performance with a low standard deviation across all runs. However, when $C$ was unknown our method dropped the most compared to ASC methods since we lost an entire cluster. SOM based approximation scores above $k$-means approximation across all similarity measures when $C$ was given, due to high quality graphs provided by SOM. For memory footprint, our method used 6.76\% of edges compared to a full graph out of all points in \texttt{iris}.

\begin{table}
	\centering
	\caption{Testing competing methods on real datasets for 100 runs. Three rows for each dataset: \texttt{ACC}: mean accuracy $\pm$ standard deviation, \texttt{ARI}: mean adjusted Rand index $\pm$ standard deviation, and \texttt{E\%}: the percentage of edges compared to edges in a fully connected graph. Bold values are the best scores.}
	\includegraphics[width=\textwidth,height=20cm,keepaspectratio]{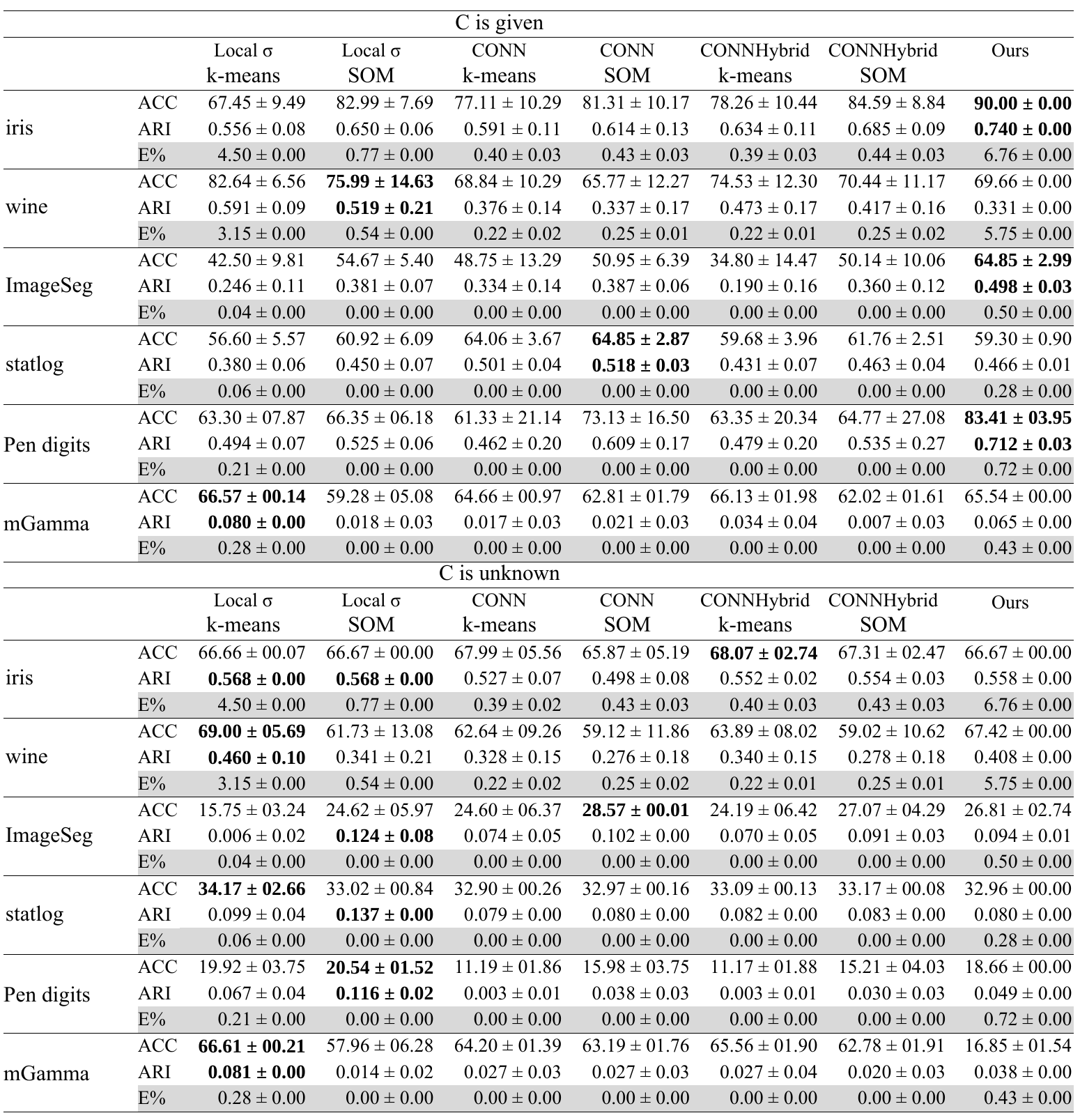}	
	\label{Table:Table-03}
\end{table}

In \texttt{wine} dataset, our method lagged behind ASC methods with a little performance drop when $C$ was unknown. We went to investigate that by projecting points in \texttt{wine} dataset onto their first three principle components. The three clusters were of a convex shape and similar density. In such cases, we do not expect our method to outperform ASC methods that are capable of capturing convex clusters. The inconsistency of ASC methods continues to persist with the standard deviation of runs reached 14\% of ACC and 0.2 of ARI.

Our method achieved the highest score for \texttt{ImageSeg} dataset when $C$ was given. Projecting the points onto first three principal components reveal non-convex clusters with varying densities. This explains the high score delivered by our method. For \texttt{statlog} dataset, the performance of our method was in line with ASC methods with a slight advantage for methods based on CONN similarity measure.

For the last two datasets: \texttt{Pen digits} and \texttt{mGamma}, the value of the baseline distribution of 7 neighbors was not providing good results. Therefore, it was set manually to be 50 neighbors after testing a range of values. This change makes our method the best performer in \texttt{Pen digits} dataset when $C$ was given. For \texttt{mGamma} dataset, the proposed method was close to the best performer when $C$ was given. But when $C$ was unknown it produced 13 clusters compared to 2 clusters in the ground truth, causing a sharp decline in ACC.

\begin{figure}
	\centering
	\includegraphics[width=\textwidth,height=20cm,keepaspectratio]{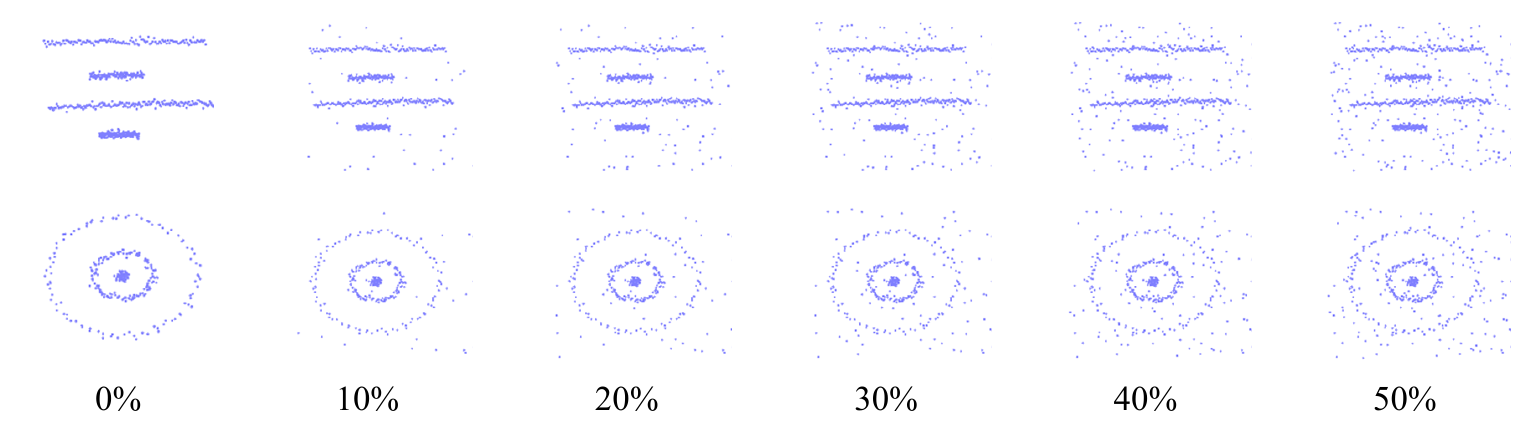}	
	\caption{Datasets used for noise robustness test contaminated by 10\% to 50\% noisy points.}
	\label{Fig:Fig-07}
\end{figure}

\begin{figure}
	\centering
	\includegraphics[width=0.95\textwidth,height=20cm,keepaspectratio]{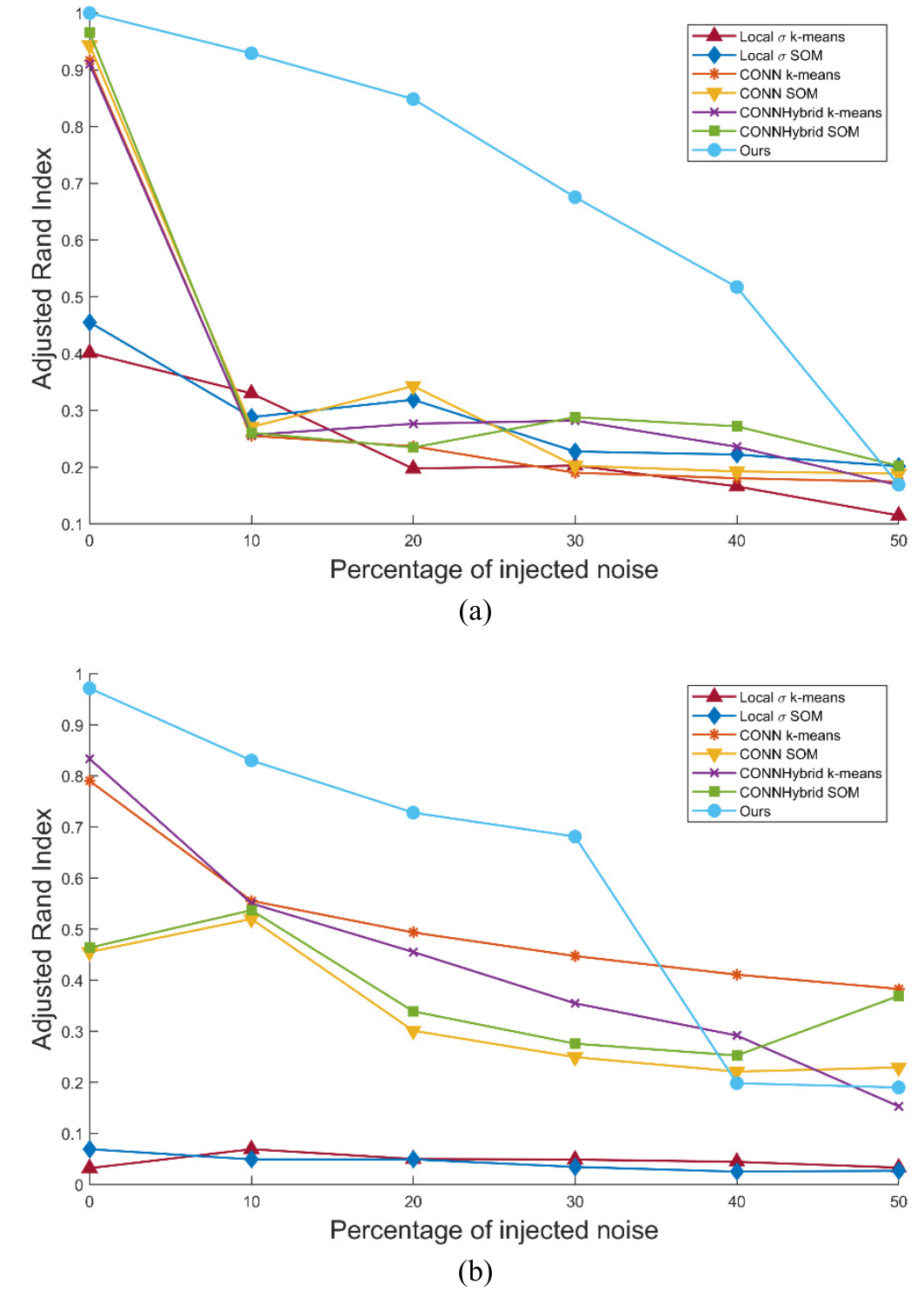}	
	\caption{Results of noise robustness test; (a) results on lines datasets; (b) results on rings dataset (best viewed in color).}
	\label{Fig:Fig-08}
\end{figure}

\subsection{Noise robustness test}
\label{Experiments-4}

The last experiment was testing competing methods’ robustness against increasing noise, where $C$ was given. The datasets in Fig.\ \ref{Fig:Fig-07} were retrieved from \cite{RN237} and the evaluation in Fig.\ \ref{Fig:Fig-08} was based on ARI. In general, CONN based similarities performed better when data was clean. With increasing noise, ASC methods start to drop in performance. On the other hand, our method continues to score better than ASC methods with noise approaching 30\% of the data. Even with 40\% noise, our method delivered the highest score on lines dataset. The core difference between our method and ASC methods is that our method cuts noisy points since they do not match the lines/rings density and ASC tries to accommodate noisy points due to its vector quantization component.

\subsection{Experiments for Integration with SpectralNet}
\label{Experiments-5}

\begin{figure}
	\centering
	\includegraphics[width=\textwidth,height=20cm,keepaspectratio]{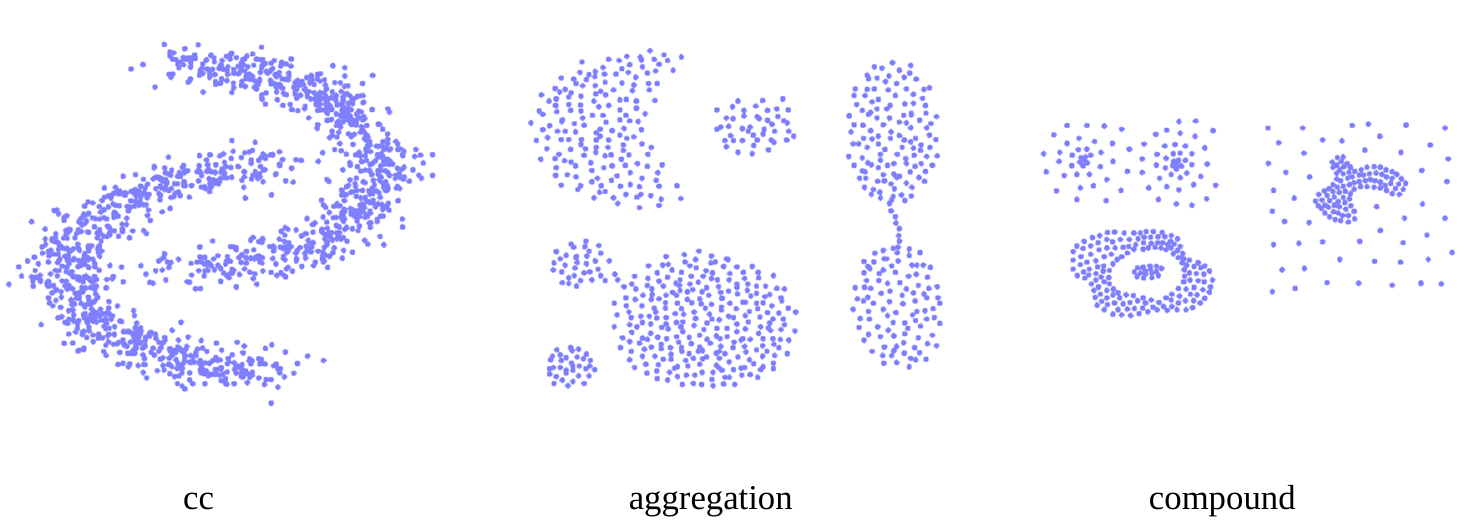}	
	\caption{Datasets used for SpectralNet experiments.}
	\label{Fig:Fig-09}
\end{figure}

For Integration with spectral clustering using deep neural networks (SpectralNet), we used three synthetic datasets, three methods, four evaluation metrics. The three datasets are: \texttt{cc}, \texttt{aggregation}, and \texttt{compound}, shown in Fig.\ \ref{Fig:Fig-09}. Two of the three methods were designed as described by \citet{RN360}, where each data point is paired with $k$ of its nearest neighbors to form positive points. In our experiments we set $k$ as $k=2$ and $k=4$. Once positive pairs are constructed, an equal number of randomly selected farther neighbors is set as negative pairs. In the third method we let the proposed method detailed in Algorithm \ref{Alg:Alg-01} to decide the number of positive pairs. Then, an equal number of farther neighbors as set as negative pairs. For evaluation metrics, we used clustering accuracy and ARI that are described in equations \ref{Eq-008} and \ref{Eq-009} respectively. In addition to ACC and ARI we used normalized mutual information (NMI) as an evaluation metric because it was reported in the original SpectralNet paper \cite{RN360}. NMI is defined as: 
\begin{equation}
	NMI(T,L)=\frac{I(T;L)}{max\{H(T),H(L)\}}\ ,
	\label{Eq-010}
\end{equation}
where $T$ and $L$ be ground truth labels and labels obtained through clustering respectively. $I(T;L)$ denotes the mutual information between $T$ and $L$, and $H(\cdot)$ denotes their entropy. We also used the total number of pairs as an indicator of computational efficiency.

Table \ref{Table:Table-04} displays the results of our SpectralNet experiments. In \texttt{cc} dataset, our proposed method outperformed the method with fixed $k$. Our method achieved ARI score of 0.800 with an increase of 30\% over its closest competitor. This performance was largely due to the higher number of pairs passed to Siamese net. Our method passed 21,478 pairs on average, compared to 6,000 and 12,000 pairs passed by $k=2$ and $k=4$ respectively. In \texttt{aggregation} dataset, $k=2$ was the best performer with ARI score of 0.808. Our method scored lower ARI than $k=2$, close to 10\%. However, the performance of $k=4$ was worse than ours, 40\% when compared to ground truth. In \texttt{compound} dataset, our method was the best performer with ARI score of 0.705 with a difference of 15\% from the second performer.

This experiment reveals that although Siamese nets are superior in building an affinity matrix, they still need an informative selection of positive and negative pairs. Fixing the number of positive pairs for all points limits the influence of points in dense regions. Points in dense regions should be able to pair with more neighbors to strengthen intra-cluster connections.

\begin{table}
	\centering
	\caption{Results of experiments for Integration with SpectralNet for 10 runs. Four rows for each dataset: \texttt{ACC}: mean accuracy $\pm$ standard deviation, \texttt{ARI}: mean adjusted Rand index $\pm$ standard deviation, \texttt{NMI}: mean normalized mutual information $\pm$ standard deviation, and \texttt{Total pairs}: the total number of positive and negative pairs passed to Siamese net. Bold values are the best scores.}
	\includegraphics[width=0.7\textwidth,height=20cm,keepaspectratio]{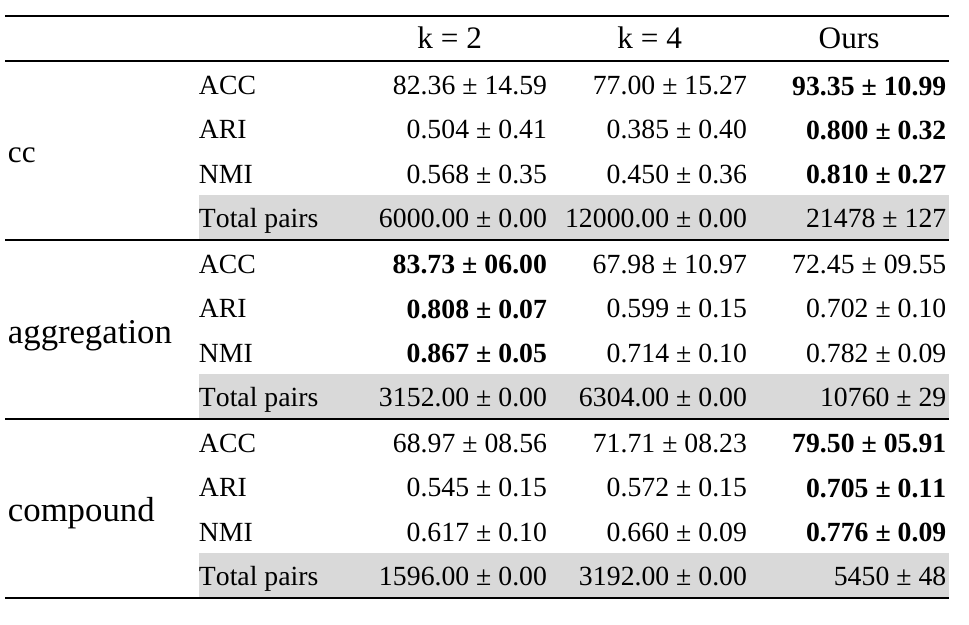}	
	\label{Table:Table-04}
\end{table}

\section{Conclusion}
\label{Conclusion}

Spectral clustering is a clustering paradigm with heavy computational demands. These demands were closely studied in the literature, and many solutions were proposed. However, many of these solutions compromised on consistency of clustering, or local statistics that are crucial for sparseness and noise detection. We proposed a series of refinement stages to the well-known $k$-nearest neighbor graph. The obtained graph can detect sparse clusters and noisy points. Also, our method was capable of delivering consistent clustering over multiple runs, since it was based on $k$-nearest neighbor graph with no random operations involved. Compared to approximate spectral clustering (ASC), the proposed method detected sparse clusters, and achieved accurate clustering with substantial noise.

The future directions of this work lie in two areas: 1) improving memory efficiency, and 2) automatically finding number of neighbors included in the baseline distribution. For memory efficiency, our method still requires memory larger than ASC methods to store the graph. ASC benefits from reducing nodes through vector quantization. Reducing nodes in a deterministic way would improve the memory efficiency of this work. Additionally, the number of neighbors included in the baseline distribution was tuned manually. The baseline distribution was used as a reference to whether or not further edges will be eliminated.





\begin{singlespace}
\bibliographystyle{elsarticle-num-names}
\bibliography{mybibfile}

\begin{thebibliography}{41}
\expandafter\ifx\csname natexlab\endcsname\relax\def\natexlab#1{#1}\fi
\providecommand{\url}[1]{\texttt{#1}}
\providecommand{\href}[2]{#2}
\providecommand{\path}[1]{#1}
\providecommand{\DOIprefix}{doi:}
\providecommand{\ArXivprefix}{arXiv:}
\providecommand{\URLprefix}{URL: }
\providecommand{\Pubmedprefix}{pmid:}
\providecommand{\doi}[1]{\href{http://dx.doi.org/#1}{\path{#1}}}
\providecommand{\Pubmed}[1]{\href{pmid:#1}{\path{#1}}}
\providecommand{\bibinfo}[2]{#2}
\ifx\xfnm\relax \def\xfnm[#1]{\unskip,\space#1}\fi
\bibitem[{Shi and Malik(2000)}]{RN261}
\bibinfo{author}{J.~Shi}, \bibinfo{author}{J.~Malik},
\newblock \bibinfo{title}{Normalized cuts and image segmentation},
\newblock \bibinfo{journal}{IEEE Transactions on Pattern Analysis and Machine
  Intelligence} \bibinfo{volume}{22} (\bibinfo{year}{2000})
  \bibinfo{pages}{888--905}. \DOIprefix\doi{10.1109/34.868688}.
\bibitem[{Ng et~al.(2002)Ng, Jordan, and Weiss}]{RN233}
\bibinfo{author}{A.~Y. Ng}, \bibinfo{author}{M.~I. Jordan},
  \bibinfo{author}{Y.~Weiss},
\newblock \bibinfo{title}{On spectral clustering: Analysis and an algorithm},
\newblock \bibinfo{journal}{Advances in Neural Information Processing Systems}
  (\bibinfo{year}{2002}).
\bibitem[{von Luxburg(2007)}]{RN234}
\bibinfo{author}{U.~von Luxburg},
\newblock \bibinfo{title}{A tutorial on spectral clustering},
\newblock \bibinfo{journal}{Statistics and Computing} \bibinfo{volume}{17}
  (\bibinfo{year}{2007}) \bibinfo{pages}{395--416}.
  \DOIprefix\doi{10.1007/s11222-007-9033-z}.
\bibitem[{Filippone et~al.(2008)Filippone, Camastra, Masulli, and
  Rovetta}]{RN295}
\bibinfo{author}{M.~Filippone}, \bibinfo{author}{F.~Camastra},
  \bibinfo{author}{F.~Masulli}, \bibinfo{author}{S.~Rovetta},
\newblock \bibinfo{title}{A survey of kernel and spectral methods for
  clustering},
\newblock \bibinfo{journal}{Pattern Recognition} \bibinfo{volume}{41}
  (\bibinfo{year}{2008}) \bibinfo{pages}{176--190}. \URLprefix
  \url{http://www.sciencedirect.com/science/article/pii/S0031320307002580}.
  \DOIprefix\doi{https://doi.org/10.1016/j.patcog.2007.05.018}.
\bibitem[{Wang and Dong(2012)}]{RN228}
\bibinfo{author}{L.~Wang}, \bibinfo{author}{M.~Dong},
\newblock \bibinfo{title}{Multi-level low-rank approximation-based spectral
  clustering for image segmentation},
\newblock \bibinfo{journal}{Pattern Recognition Letters} \bibinfo{volume}{33}
  (\bibinfo{year}{2012}) \bibinfo{pages}{2206--2215}.
  \DOIprefix\doi{http://dx.doi.org/10.1016/j.patrec.2012.07.024}.
\bibitem[{Alshammari and Takatsuka(2019)}]{RN296}
\bibinfo{author}{M.~Alshammari}, \bibinfo{author}{M.~Takatsuka},
\newblock \bibinfo{title}{Approximate spectral clustering with eigenvector
  selection and self-tuned k},
\newblock \bibinfo{journal}{Pattern Recognition Letters} \bibinfo{volume}{122}
  (\bibinfo{year}{2019}) \bibinfo{pages}{31--37}. \URLprefix
  \url{http://www.sciencedirect.com/science/article/pii/S0167865519300315}.
  \DOIprefix\doi{https://doi.org/10.1016/j.patrec.2019.02.006}.
\bibitem[{Tasdemir(2013)}]{RN297}
\bibinfo{author}{K.~Tasdemir},
\newblock \bibinfo{title}{A hybrid similarity measure for approximate spectral
  clustering of remote sensing images},
\newblock \bibinfo{journal}{2013 IEEE International Geoscience and Remote
  Sensing Symposium - IGARSS}  (\bibinfo{year}{2013})
  \bibinfo{pages}{3136--3139}. \DOIprefix\doi{10.1109/IGARSS.2013.6723491}.
\bibitem[{Tasdemir et~al.(2015)Tasdemir, Yalcin, and Yildirim}]{RN276}
\bibinfo{author}{K.~Tasdemir}, \bibinfo{author}{B.~Yalcin},
  \bibinfo{author}{I.~Yildirim},
\newblock \bibinfo{title}{Approximate spectral clustering with utilized
  similarity information using geodesic based hybrid distance measures},
\newblock \bibinfo{journal}{Pattern Recognition} \bibinfo{volume}{48}
  (\bibinfo{year}{2015}) \bibinfo{pages}{1465--1477}.
  \DOIprefix\doi{https://doi.org/10.1016/j.patcog.2014.10.023}.
\bibitem[{Tyuryukanov et~al.(2018)Tyuryukanov, Popov, Meijden, and
  Terzija}]{RN288}
\bibinfo{author}{I.~Tyuryukanov}, \bibinfo{author}{M.~Popov},
  \bibinfo{author}{M.~v.~d. Meijden}, \bibinfo{author}{V.~Terzija},
\newblock \bibinfo{title}{Discovering clusters in power networks from
  orthogonal structure of spectral embedding},
\newblock \bibinfo{journal}{IEEE Transactions on Power Systems}
  (\bibinfo{year}{2018}). \DOIprefix\doi{10.1109/TPWRS.2018.2854962}.
\bibitem[{Wang et~al.(2017)Wang, Lin, and Wang}]{RN292}
\bibinfo{author}{T.~Wang}, \bibinfo{author}{H.~Lin}, \bibinfo{author}{P.~Wang},
\newblock \bibinfo{title}{Weighted-spectral clustering algorithm for detecting
  community structures in complex networks},
\newblock \bibinfo{journal}{Artificial Intelligence Review}
  \bibinfo{volume}{47} (\bibinfo{year}{2017}) \bibinfo{pages}{463--483}.
  \URLprefix \url{https://doi.org/10.1007/s10462-016-9488-4}.
  \DOIprefix\doi{10.1007/s10462-016-9488-4}.
\bibitem[{Kohonen(2013)}]{RN298}
\bibinfo{author}{T.~Kohonen},
\newblock \bibinfo{title}{Essentials of the self-organizing map},
\newblock \bibinfo{journal}{Neural Networks} \bibinfo{volume}{37}
  (\bibinfo{year}{2013}) \bibinfo{pages}{52--65}. \URLprefix
  \url{http://www.sciencedirect.com/science/article/pii/S0893608012002596}.
  \DOIprefix\doi{https://doi.org/10.1016/j.neunet.2012.09.018}.
\bibitem[{Arthur and Vassilvitskii(2007)}]{RN240}
\bibinfo{author}{D.~Arthur}, \bibinfo{author}{S.~Vassilvitskii},
\newblock \bibinfo{title}{K-means++: The advantages of careful seeding},
\newblock \bibinfo{journal}{Proceedings of the Annual ACM-SIAM Symposium on
  Discrete Algorithms} \bibinfo{volume}{07-09-January-2007}
  (\bibinfo{year}{2007}) \bibinfo{pages}{1027--1035}.
\bibitem[{Yan et~al.(2009)Yan, Huang, and Jordan}]{RN232}
\bibinfo{author}{D.~Yan}, \bibinfo{author}{L.~Huang}, \bibinfo{author}{M.~I.
  Jordan},
\newblock \bibinfo{title}{Fast approximate spectral clustering},
\newblock \bibinfo{journal}{Proceedings of the 15th ACM SIGKDD international
  conference on Knowledge discovery and data mining}  (\bibinfo{year}{2009})
  \bibinfo{pages}{907--916}.
\bibitem[{Tasdemir(2012)}]{RN275}
\bibinfo{author}{K.~Tasdemir},
\newblock \bibinfo{title}{Vector quantization based approximate spectral
  clustering of large datasets},
\newblock \bibinfo{journal}{Pattern Recognition} \bibinfo{volume}{45}
  (\bibinfo{year}{2012}) \bibinfo{pages}{3034--3044}.
  \DOIprefix\doi{https://doi.org/10.1016/j.patcog.2012.02.012}.
\bibitem[{Alshammari and Takatsuka(2018)}]{RN285}
\bibinfo{author}{M.~Alshammari}, \bibinfo{author}{M.~Takatsuka},
\newblock \bibinfo{title}{Approximate spectral clustering using topology
  preserving methods and local scaling},
\newblock \bibinfo{journal}{The 25th International Conference on Neural
  Information Processing (ICONIP 2018)}  (\bibinfo{year}{2018}).
\bibitem[{Zelnik-Manor and Perona(2005)}]{RN237}
\bibinfo{author}{L.~Zelnik-Manor}, \bibinfo{author}{P.~Perona},
\newblock \bibinfo{title}{Self-tuning spectral clustering},
\newblock \bibinfo{journal}{Advances in Neural Information Processing Systems}
  (\bibinfo{year}{2005}) \bibinfo{pages}{1601--1608}.
\bibitem[{Shi and Malik(1997)}]{RN362}
\bibinfo{author}{J.~Shi}, \bibinfo{author}{J.~Malik},
\newblock \bibinfo{title}{Normalized cuts and image segmentation},
\newblock \bibinfo{year}{1997}, pp. \bibinfo{pages}{731--737}. \URLprefix
  \url{https://www.scopus.com/inward/record.uri?eid=2-s2.0-0030646927&partnerID=40&md5=42573d155d149f8aaa0e35a0ea7664a2}.
  \DOIprefix\doi{Doi 10.1109/Cvpr.1997.609407}.
\bibitem[{Meila and Shi(2001)}]{RN279}
\bibinfo{author}{M.~Meila}, \bibinfo{author}{J.~Shi},
\newblock \bibinfo{title}{Learning segmentation by random walks},
\newblock \bibinfo{journal}{Advances in Neural Information Processing Systems}
  (\bibinfo{year}{2001}).
\bibitem[{Sugiyama(2007)}]{RN274}
\bibinfo{author}{M.~Sugiyama},
\newblock \bibinfo{title}{Dimensionality reduction of multimodal labeled data
  by local fisher discriminant analysis},
\newblock \bibinfo{journal}{Journal of machine learning research}
  \bibinfo{volume}{8} (\bibinfo{year}{2007}) \bibinfo{pages}{1027--1061}.
\bibitem[{Lloyd(1982)}]{RN309}
\bibinfo{author}{S.~Lloyd},
\newblock \bibinfo{title}{Least squares quantization in pcm},
\newblock \bibinfo{journal}{IEEE Transactions on Information Theory}
  \bibinfo{volume}{28} (\bibinfo{year}{1982}) \bibinfo{pages}{129--137}.
  \DOIprefix\doi{10.1109/TIT.1982.1056489}.
\bibitem[{Kohonen(1990)}]{RN42}
\bibinfo{author}{T.~Kohonen},
\newblock \bibinfo{title}{The self-organizing map},
\newblock \bibinfo{journal}{Proceedings of the IEEE} \bibinfo{volume}{78}
  (\bibinfo{year}{1990}) \bibinfo{pages}{1464--1480}.
\bibitem[{Martinetz and Schulten(1991)}]{RN259}
\bibinfo{author}{T.~Martinetz}, \bibinfo{author}{K.~Schulten},
\newblock \bibinfo{title}{A ``neural-gas'' network learns topologies},
\newblock \bibinfo{journal}{University of Illinois at Urbana-Champaign
  Champaign, IL}  (\bibinfo{year}{1991}).
\bibitem[{Martinetz and Schulten(1994)}]{RN257}
\bibinfo{author}{T.~Martinetz}, \bibinfo{author}{K.~Schulten},
\newblock \bibinfo{title}{Topology representing networks},
\newblock \bibinfo{journal}{Neural Networks} \bibinfo{volume}{7}
  (\bibinfo{year}{1994}) \bibinfo{pages}{507--522}.
  \DOIprefix\doi{10.1016/0893-6080(94)90109-0}.
\bibitem[{Wang et~al.(2017)Wang, Nie, and Yu}]{RN359}
\bibinfo{author}{R.~Wang}, \bibinfo{author}{F.~Nie}, \bibinfo{author}{W.~Yu},
\newblock \bibinfo{title}{Fast spectral clustering with anchor graph for large
  hyperspectral images},
\newblock \bibinfo{journal}{IEEE Geoscience and Remote Sensing Letters}
  \bibinfo{volume}{14} (\bibinfo{year}{2017}) \bibinfo{pages}{2003--2007}.
  \DOIprefix\doi{10.1109/LGRS.2017.2746625}.
\bibitem[{Guo and Ye(2019)}]{RN358}
\bibinfo{author}{J.~Guo}, \bibinfo{author}{J.~Ye},
\newblock \bibinfo{title}{Anchors bring ease: An embarrassingly simple approach
  to partial multi-view clustering},
\newblock \bibinfo{journal}{Proceedings of the AAAI Conference on Artificial
  Intelligence} \bibinfo{volume}{33} (\bibinfo{year}{2019})
  \bibinfo{pages}{118--125}.
\bibitem[{Correa and Lindstrom(2012)}]{RN254}
\bibinfo{author}{C.~D. Correa}, \bibinfo{author}{P.~Lindstrom},
\newblock \bibinfo{title}{Locally-scaled spectral clustering using empty region
  graphs},
\newblock in: \bibinfo{booktitle}{Proceedings of the 18th ACM SIGKDD
  International Conference on Knowledge Discovery and Data Mining}, KDD ’12,
  \bibinfo{publisher}{Association for Computing Machinery},
  \bibinfo{address}{New York, NY, USA}, \bibinfo{year}{2012}, p.
  \bibinfo{pages}{1330–1338}. \URLprefix
  \url{https://doi.org/10.1145/2339530.2339736}.
  \DOIprefix\doi{10.1145/2339530.2339736}.
\bibitem[{İnkaya et~al.(2015)İnkaya, Kayalıgil, and Özdemirel}]{RN303}
\bibinfo{author}{T.~İnkaya}, \bibinfo{author}{S.~Kayalıgil},
  \bibinfo{author}{N.~E. Özdemirel},
\newblock \bibinfo{title}{An adaptive neighbourhood construction algorithm
  based on density and connectivity},
\newblock \bibinfo{journal}{Pattern Recognition Letters} \bibinfo{volume}{52}
  (\bibinfo{year}{2015}) \bibinfo{pages}{17--24}. \URLprefix
  \url{http://www.sciencedirect.com/science/article/pii/S0167865514002815}.
  \DOIprefix\doi{https://doi.org/10.1016/j.patrec.2014.09.007}.
\bibitem[{Inkaya(2015)}]{RN253}
\bibinfo{author}{T.~Inkaya},
\newblock \bibinfo{title}{A parameter-free similarity graph for spectral
  clustering},
\newblock \bibinfo{journal}{Expert Systems with Applications}
  \bibinfo{volume}{42} (\bibinfo{year}{2015}) \bibinfo{pages}{9489--9498}.
  \URLprefix
  \url{http://www.sciencedirect.com/science/article/pii/S0957417415005345}.
  \DOIprefix\doi{https://doi.org/10.1016/j.eswa.2015.07.074}.
\bibitem[{Zhengdong and Carreira-Perpinan(2008)}]{RN300}
\bibinfo{author}{L.~Zhengdong}, \bibinfo{author}{M.~A. Carreira-Perpinan},
\newblock \bibinfo{title}{Constrained spectral clustering through affinity
  propagation},
\newblock \bibinfo{journal}{2008 IEEE Conference on Computer Vision and Pattern
  Recognition}  (\bibinfo{year}{2008}) \bibinfo{pages}{1--8}.
  \DOIprefix\doi{10.1109/CVPR.2008.4587451}.
\bibitem[{Li et~al.(2009)Li, Liu, and Tang}]{RN302}
\bibinfo{author}{Z.~Li}, \bibinfo{author}{J.~Liu}, \bibinfo{author}{X.~Tang},
\newblock \bibinfo{title}{Constrained clustering via spectral regularization},
\newblock \bibinfo{journal}{2009 IEEE Conference on Computer Vision and Pattern
  Recognition}  (\bibinfo{year}{2009}) \bibinfo{pages}{421--428}.
  \DOIprefix\doi{10.1109/CVPR.2009.5206852}.
\bibitem[{Zhang et~al.(2017)Zhang, Nie, and Li}]{RN301}
\bibinfo{author}{R.~Zhang}, \bibinfo{author}{F.~Nie}, \bibinfo{author}{X.~Li},
\newblock \bibinfo{title}{Self-weighted spectral clustering with parameter-free
  constraint},
\newblock \bibinfo{journal}{Neurocomputing} \bibinfo{volume}{241}
  (\bibinfo{year}{2017}) \bibinfo{pages}{164--170}. \URLprefix
  \url{http://www.sciencedirect.com/science/article/pii/S0925231217303089}.
  \DOIprefix\doi{https://doi.org/10.1016/j.neucom.2017.01.085}.
\bibitem[{Liu et~al.(2013)Liu, Lin, Yan, Sun, Yu, and Ma}]{RN168}
\bibinfo{author}{G.~Liu}, \bibinfo{author}{Z.~Lin}, \bibinfo{author}{S.~Yan},
  \bibinfo{author}{J.~Sun}, \bibinfo{author}{Y.~Yu}, \bibinfo{author}{Y.~Ma},
\newblock \bibinfo{title}{Robust recovery of subspace structures by low-rank
  representation},
\newblock \bibinfo{journal}{IEEE Transactions on Pattern Analysis and Machine
  Intelligence} \bibinfo{volume}{35} (\bibinfo{year}{2013})
  \bibinfo{pages}{171--184}. \DOIprefix\doi{10.1109/TPAMI.2012.88}.
\bibitem[{Shaham et~al.(2018)Shaham, Stanton, Li, Nadler, Basri, and
  Kluger}]{RN360}
\bibinfo{author}{U.~Shaham}, \bibinfo{author}{K.~Stanton},
  \bibinfo{author}{H.~Li}, \bibinfo{author}{B.~Nadler},
  \bibinfo{author}{R.~Basri}, \bibinfo{author}{Y.~Kluger},
\newblock \bibinfo{title}{Spectralnet: Spectral clustering using deep neural
  networks},
\newblock \bibinfo{journal}{6th International Conference on Learning
  Representations, ICLR 2018 - Conference Track Proceedings}
  (\bibinfo{year}{2018}). \URLprefix
  \url{https://www.scopus.com/inward/record.uri?eid=2-s2.0-85071002796&partnerID=40&md5=50cd7687ec9a8a763991553362caa86c}.
\bibitem[{Bengio et~al.(2004)Bengio, Paiement, Vincent, Delalleau, Le~Roux, and
  Ouimet}]{Bengio2004OOSE}
\bibinfo{author}{Y.~Bengio}, \bibinfo{author}{J.-F. Paiement},
  \bibinfo{author}{P.~Vincent}, \bibinfo{author}{O.~Delalleau},
  \bibinfo{author}{N.~Le~Roux}, \bibinfo{author}{M.~Ouimet},
\newblock \bibinfo{title}{Out-of-sample extensions for lle, isomap, mds,
  eigenmaps, and spectral clustering},
\newblock \bibinfo{year}{2004}. \URLprefix
  \url{https://www.scopus.com/inward/record.uri?eid=2-s2.0-33947233031&partnerID=40&md5=0277a925a2ef56079a83a8f33b5b4cde}.
\bibitem[{Fowlkes et~al.(2004)Fowlkes, Belongie, Chung, and Malik}]{RN227}
\bibinfo{author}{C.~Fowlkes}, \bibinfo{author}{S.~Belongie},
  \bibinfo{author}{F.~Chung}, \bibinfo{author}{J.~Malik},
\newblock \bibinfo{title}{Spectral grouping using the nystrom method},
\newblock \bibinfo{journal}{IEEE Transactions on Pattern Analysis and Machine
  Intelligence} \bibinfo{volume}{26} (\bibinfo{year}{2004})
  \bibinfo{pages}{214--225}. \DOIprefix\doi{10.1109/TPAMI.2004.1262185}.
\bibitem[{Coifman and Lafon(2006)}]{Coifman2006Geometric}
\bibinfo{author}{R.~Coifman}, \bibinfo{author}{S.~Lafon},
\newblock \bibinfo{title}{Geometric harmonics: A novel tool for multiscale
  out-of-sample extension of empirical functions},
\newblock \bibinfo{journal}{Applied and Computational Harmonic Analysis}
  \bibinfo{volume}{21} (\bibinfo{year}{2006}) \bibinfo{pages}{31--52}.
  \URLprefix
  \url{https://www.scopus.com/inward/record.uri?eid=2-s2.0-33745398604&doi=10.1016%2fj.acha.2005.07.005&partnerID=40&md5=9996d6b162cca026a4caf1805004eff5}.
  \DOIprefix\doi{10.1016/j.acha.2005.07.005}.
\bibitem[{Hadsell et~al.(2006)Hadsell, Chopra, and
  LeCun}]{Hadsell2006Dimensionality}
\bibinfo{author}{R.~Hadsell}, \bibinfo{author}{S.~Chopra},
  \bibinfo{author}{Y.~LeCun},
\newblock \bibinfo{title}{Dimensionality reduction by learning an invariant
  mapping},
\newblock volume~\bibinfo{volume}{2}, \bibinfo{year}{2006}, pp.
  \bibinfo{pages}{1735--1742}. \URLprefix
  \url{https://www.scopus.com/inward/record.uri?eid=2-s2.0-33845594569&doi=10.1109%2fCVPR.2006.100&partnerID=40&md5=6b88a96eb9703c28939e5534c7ba6bd8}.
  \DOIprefix\doi{10.1109/CVPR.2006.100}.
\bibitem[{Shaham and Lederman(2018)}]{Shaham2018Learning}
\bibinfo{author}{U.~Shaham}, \bibinfo{author}{R.~Lederman},
\newblock \bibinfo{title}{Learning by coincidence: Siamese networks and common
  variable learning},
\newblock \bibinfo{journal}{Pattern Recognition} \bibinfo{volume}{74}
  (\bibinfo{year}{2018}) \bibinfo{pages}{52--63}. \URLprefix
  \url{https://www.scopus.com/inward/record.uri?eid=2-s2.0-85032304883&doi=10.1016%2fj.patcog.2017.09.015&partnerID=40&md5=358ba20ecd4f9a87026ad0474790aeca}.
  \DOIprefix\doi{10.1016/j.patcog.2017.09.015}.
\bibitem[{Cai et~al.(2005)Cai, He, and Han}]{RN238}
\bibinfo{author}{D.~Cai}, \bibinfo{author}{X.~He}, \bibinfo{author}{J.~Han},
\newblock \bibinfo{title}{Document clustering using locality preserving
  indexing},
\newblock \bibinfo{journal}{IEEE Transactions on Knowledge and Data
  Engineering} \bibinfo{volume}{17} (\bibinfo{year}{2005})
  \bibinfo{pages}{1624--1637}. \DOIprefix\doi{10.1109/TKDE.2005.198}.
\bibitem[{Hubert and Arabie(1985)}]{RN365}
\bibinfo{author}{L.~Hubert}, \bibinfo{author}{P.~Arabie},
\newblock \bibinfo{title}{Comparing partitions},
\newblock \bibinfo{journal}{Journal of Classification} \bibinfo{volume}{2}
  (\bibinfo{year}{1985}) \bibinfo{pages}{193--218}. \URLprefix
  \url{https://www.scopus.com/inward/record.uri?eid=2-s2.0-0000008146&doi=10.1007%2fBF01908075&partnerID=40&md5=bd03cf70caee7de0ccf3c0dd431b97ca}.
  \DOIprefix\doi{10.1007/BF01908075}.
\bibitem[{Rand(1971)}]{RN364}
\bibinfo{author}{W.~M. Rand},
\newblock \bibinfo{title}{Objective criteria for the evaluation of clustering
  methods},
\newblock \bibinfo{journal}{Journal of the American Statistical Association}
  \bibinfo{volume}{66} (\bibinfo{year}{1971}) \bibinfo{pages}{846--850}.
  \URLprefix
  \url{https://www.scopus.com/inward/record.uri?eid=2-s2.0-84950632109&doi=10.1080%2f01621459.1971.10482356&partnerID=40&md5=3bcb39c5cbd4ccf7dec3e0fa080b1759}.
  \DOIprefix\doi{10.1080/01621459.1971.10482356}.

\end{thebibliography}
\end{singlespace}

\end{document}